\documentclass[utf8]{FrontiersinHarvard} % for articles in journals using the Harvard Referencing Style (Author-Date), for Frontiers Reference Styles by Journal: https://zendesk.frontiersin.org/hc/en-us/articles/360017860337-Frontiers-Reference-Styles-by-Journal
\usepackage{url,hyperref,lineno,microtype,subcaption}
\usepackage[onehalfspacing]{setspace}
\usepackage{amsmath}
\usepackage{booktabs}
\usepackage{multirow}
\usepackage{etoolbox}
\makeatletter
\patchcmd{\maketitle}{\vskip 0.5em}{\vskip -0.5em}{}{}
\makeatother
\usepackage{hyperref}
\usepackage{url}

\usepackage{graphicx}
\usepackage{multirow}
\usepackage{natbib}

\usepackage{tcolorbox}
\usepackage{listings}
\usepackage[utf8]{inputenc}
\usepackage[T1]{fontenc}
\usepackage{multirow}
\usepackage{etoolbox}
\makeatletter
\patchcmd{\maketitle}{\vskip 0.5em}{\vskip -0.5em}{}{}
\makeatother
\usepackage{hyperref}
\usepackage{url}

\def\keyFont{\fontsize{8}{11}\helveticabold}
\def\firstAuthorLast{Cooray {et~al.}} % use et al. only if more than 1 author

\def\Authors{
\textbf{Lakshan Cooray\,$^{1,*}$, 
Deshan Sumanathilaka\,$^{2}$, 
Pattigadapa Venkatesh Raju\,$^{3}$}
}

\def\Address{
$^{1}$\textit{School of Computing, Informatics Institute of Technology, Colombo 06, Western Province, Sri Lanka}\\
$^{2}$\textit{School of Mathematics and Computer Science, Swansea University, Swansea, SA1
8EN, UK}\\
$^{3}$\textit{R\&D, Zame AI}
}

\def\corrAuthor{Lakshan Cooray}
\def\corrEmail{lakshan.20221470@iit.ac.lk}

\begin{document}
\onecolumn
\firstpage{1}

\title[SLMs for Context-Summarized Multi-Turn CS QA]{Can Small Language Models Handle Context-Summarized Multi-Turn Customer-Service QA? A Synthetic Data-Driven Comparative Evaluation} 

\author[\firstAuthorLast ]{\Authors} %This field will be automatically populated
\address{} %This field will be automatically populated
\correspondance{} %This field will be automatically populated

\extraAuth{}% If there are more than 1 corresponding author, comment this line and uncomment the next one.
%\extraAuth{corresponding Author2 \\ Laboratory X2, Institute X2, Department X2, Organization X2, Street X2, City X2 , State XX2 (only USA, Canada and Australia), Zip Code2, X2 Country X2, email2@uni2.edu}

\maketitle

\begin{abstract}

Customer-service question answering (QA) systems increasingly rely on conversational
language understanding. While Large Language Models (LLMs) achieve strong performance, their
high computational cost and deployment constraints limit practical use in resource-constrained
environments. Small Language Models (SLMs) provide a more efficient alternative, yet their
effectiveness for multi-turn customer-service QA remains underexplored, particularly in scenarios
requiring dialogue continuity and contextual understanding. In this study, we evaluate whether
instruction-tuned SLMs, fine-tuned using parameter-efficient finetuning, can effectively handle
context-summarized multi-turn customer-service QA while preserving contextual consistency,
response quality and task relevance under computational constraints. We further investigate
instruction-tuned SLMs for context-summarized multi-turn customer-service QA using a
history summarization strategy to preserve essential conversational state and introduce a
conversation stage-based qualitative analysis to evaluate model behavior across different phases
of customer-service interactions. The main contributions of this work include the application of
parameter-efficient fine-tuning to adapt SLMs for context-summarized multi-turn customer-service
QA, a synthetic data construction pipeline for generating a context-summarized multi-turn QA
dataset, and a structured evaluation framework combining quantitative metrics with human and
LLM-as-a-judge assessments for customer-service QA evaluation. Nine instruction-tuned SLMs are
evaluated against three commercial LLMs using lexical and semantic similarity metrics alongside
qualitative assessments, including human evaluation and LLM-as-a-judge methods. Results show
notable variation across SLMs, with some models demonstrating near-LLM performance, while
others struggle to maintain dialogue continuity and contextual alignment. These findings highlight
both the potential and current limitations of low-parameter language models for real-world
customer-service QA systems.

\tiny
\keyFont{ \section{Keywords:} Small Language Models (SLMs), Multi-turn Dialogue Systems, Context Summarization, Customer Service Question Answering (QA), Instruction-Tuned Language Models, Qualitative Analysis, Conversational Stage-based Assessment}
%All article types: you may provide up to 8 keywords; at least 5 are mandatory.
\end{abstract}

\section{Introduction}
\label{sec:introduction}

Customer service interactions are a critical component of modern business operations, directly influencing customer satisfaction, organizational reputation and operational efficiency. Customers across sectors such as banking, telecommunications and e-commerce frequently contact service providers to resolve issues, seek information, or request account modifications \citep{bird2022improvingcustomerservicechatbots}. These interactions typically involve multiple exchanges between clients and agents, incorporate domain-specific terminology and require contextual continuity across dialogue turns. Manual handling of such conversations imposes substantial operational costs related to agent recruitment, training and supervision, motivating growing interest in automation technologies.

Early customer service automation relied on rule-based systems and statistical machine learning models such as Support Vector Machines and Hidden Markov Models. Although effective for basic intent detection, these approaches struggled with linguistic variability and long-range dependencies in multi-turn dialogue (interactions consisting of multiple conversational exchanges between a client and an agent) \citep{wang-etal-2017-telecom}. Transformer architectures advanced the field by enabling contextual representations through self-attention, supporting more coherent conversations \citep{vaswani2023attentionneed}. Building on this, LLMs showed strong ability in understanding context, reasoning over queries and generating fluent customer service responses \citep{wulf2024exploringpotentiallargelanguage, chat-glm-electronic}. However, their large size leads to high computational cost, latency and dependence on cloud APIs, intensifying privacy and data governance concerns since customer interactions often contain sensitive or personally identifiable information. Such data sharing raises legal and ethical issues in regulated domains requiring strict compliance \citep{ilse2024comparative, Kamisetty2025TransformingBankingLLMs}. These factors limit deployment in resource-constrained or on-premise settings.

SLMs, typically defined as models with less than ten billion parameters, have emerged as efficient alternatives \citep{belcak2025smalllanguagemodelsfuture, meconi2025largelanguagemodelsunderstand, xu2025evaluatingsmalllanguagemodels}. Recent SLM families such as SmolLM, Qwen, Phi and LLaMA demonstrate strong instruction-following (ability to generate responses based on explicit user instructions) and reasoning capabilities while remaining deployable on standard hardware \citep{allal2025smollm2smolgoesbig}. Parameter-efficient fine-tuning methods further enable adaptation to specialized domains with reduced computational and memory requirements \citep{hu2021loralowrankadaptationlarge, dettmers2023QLoRAefficientfinetuningquantized, zhang2023adaloraadaptivebudgetallocation, li-liang-2021-prefix}, suggesting that SLMs offer a practical balance between performance and efficiency for customer-service automation.

Despite growing interest, the effectiveness of SLMs for customer-service QA remains underexplored, particularly in multi-turn client-agent interactions requiring dialogue continuity and contextual understanding across turns. Existing research has largely focused on single-turn QA settings, where conversational history is not modeled or leveraged \citep{wang-etal-2017-telecom, 11018908}. No work has systematically evaluated recently introduced instruction-tuned SLMs (SLMs fine-tuned to follow task-specific instructions) under multi-turn customer-service settings \citep{chat-glm-electronic, lijaya2025comparative}. Evaluation practices are often inconsistent, relying either on automatic metrics such as ROUGE (Recall-Oriented Understudy for Gisting Evaluation) and BERTScore (a semantic similarity metric based on contextual embeddings) \citep{lin-2004-rouge, zhang2020bertscoreevaluatingtextgeneration} or on qualitative approaches such as human assessment and LLM-as-a-judge methods (using a LLM to evaluate generated responses) in isolation \citep{liu-etal-2023-g, park2024pairevalopendomaindialogueevaluation}. Additionally, the lack of publicly available English benchmark datasets for multi-turn customer-service conversations limits experimental comparability, as existing datasets such as TelBench and TeleEval CS are restricted to Chinese and Korean languages \citep{lee-etal-2024-telbench, li2025performance}.

To address these limitations, this study systematically evaluates fine-tuned, instruction-tuned SLMs for context-summarized multi-turn customer-service QA. A synthetic data construction pipeline is introduced to mitigate the limited availability of publicly accessible context-summarized multi-turn customer-service QA data. The pipeline transforms single-turn QA instances into structured multi-turn interactions, applies context summarization to refine dialogue history and performs LLM-based response refinement prior to fine-tuning. In addition, a conversation stage-based segmentation is employed to categorize interactions into early, mid and late stages, enabling stage-wise qualitative analysis of model behavior across different phases of customer-service conversations. Performance is assessed using a comprehensive evaluation framework combining lexical and semantic similarity metrics with LLM-as-a-judge and human assessment. All models are evaluated under identical experimental conditions in context-summarized multi-turn customer-service settings, enabling a fair comparison between SLMs and LLMs.

Our main contributions are as follows:
\begin{itemize}
    \item A systematic evaluation of fine-tuned, instruction-tuned SLMs for context-summarized multi-turn customer-service QA.
    \item A synthetic data construction and fine-tuning pipeline that integrates multi-turn context summarization with LLM-based response refinement to produce privacy-preserving training data for SLMs.
    \item  A comparative evaluation of fine-tuned SLMs against state-of-the-art LLMs for context-summarized multi-turn customer-service QA using automatic metrics, human evaluation and conversation stage-based analysis.
\end{itemize}

This paper is structured as follows: Section~\ref{sec:related} reviews related work and Section~\ref{sec:method} details the proposed methodology and dataset construction. Section~\ref{sec:experiments} presents the experimental setup, while Section~\ref{subsec:obtained_results} discusses the obtained results. Section~\ref{sec:conclusion} concludes the paper and outlines future research directions.

\section{Related Work}
\label{sec:related}

\subsection{Usage of NLP in Customer Service QA}

\begin{figure*}[t]
    \centering
    \includegraphics[width=0.95\textwidth, height=6.5cm, keepaspectratio]{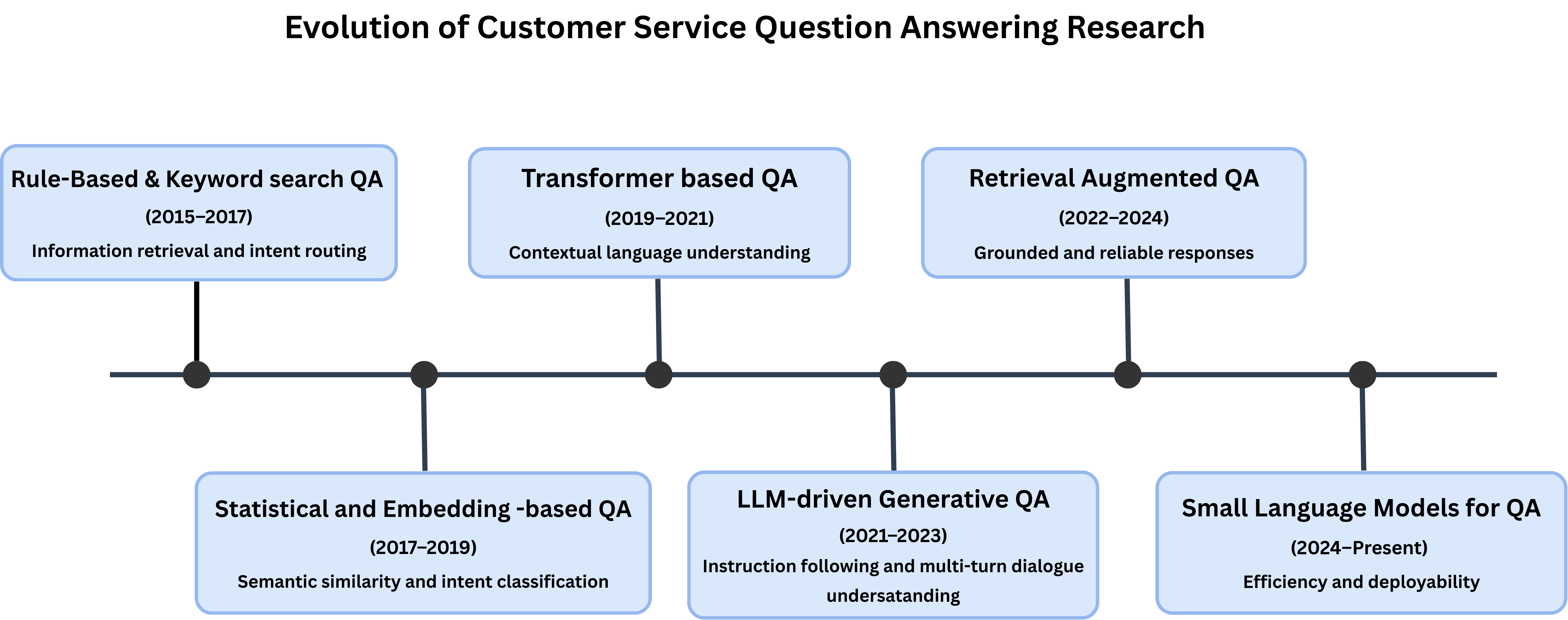}
    \caption{Overview of the evolution of customer-service question answering research.}
    \label{fig:evolutionQA}
\end{figure*}

Early customer-service QA systems relied on retrieval and embedding-based methods. For example, \citep{wang-etal-2017-telecom} proposed a hybrid system combining BM25 keyword search, a probabilistic ranking algorithm based on term frequency and document relevance with Word2Vec and Doc2Vec embeddings, enhanced by a k-nearest neighbor classifier for intent awareness and answer re-ranking. While effective in structured settings, these approaches lacked advanced contextual understanding and struggled to maintain dialogue continuity across multiple exchanges. Subsequent research introduced progressively more advanced paradigms, evolving from retrieval-based systems to transformer models, LLMs and more recently SLMs. This overall progression of customer-service QA research is illustrated in Figure~\ref{fig:evolutionQA}.

The introduction of the transformer architecture made a significant advance in conversational modeling \citep{vaswani2023attentionneed}. Pre-trained sequence-to-sequence models trained on large-scale datasets, such as customer support tweets, were adapted to domain-specific chatbots and deployed on social robots like Temi and Pepper \citep{bird2022improvingcustomerservicechatbots}. Encoder-decoder models such as T5 \citep{raffel2023exploringlimitstransferlearning} and Flan-T5 \citep{chung2022scalinginstructionfinetunedlanguagemodels} both introduced by Google further improved robustness to diverse queries through fine-tuning. More recent work showed that retrieval-augmented generation approaches using models such as LLaMA (by Meta), Gemma (by Google) and Mistral (by Mistral AI) achieved higher accuracy on reworded questions but incurred slower inference and increased system complexity \citep{11018908}.

As LLMs became dominant, research increasingly focused on efficient adaptation strategies. \citet{ilse2024comparative} compared full fine-tuning, LoRA-based parameter-efficient tuning (Low-Rank Adaptation) and domain-adaptive pre-training on models such as GPT-4, Gemini and LLaMA-2 (Meta). Domain-adaptive pre-training has obtained the strongest performance, while LoRA enabled faster and more resource-efficient adaptation, underscoring the importance of parameter-efficient fine-tuning for real-time customer-service deployment.

Applications of LLMs in customer-service QA reveal both strengths and limitations. A Swiss telecom study showed that GPT-4 could draft email responses but struggled with multi-step reasoning and hallucinations \citep{wulf2024exploringpotentiallargelanguage}. A hybrid system based on ChatGLM2-6B with LangChain, LoRA fine-tuning and reinforcement learning via Proximal Policy Optimization achieved a 74.8\% user acceptance rate, outperforming GPT-4 and baseline models \citep{chat-glm-electronic}. Emotion-aware QA systems further improved response relevance and positivity \citep{kdmile}. Despite these advances, LLM-based systems continue to face challenges related to scalability, latency and deployment cost.

\vspace{-20pt}

Reliability has been addressed through validation pipelines such as CHOPS (CHat with custOmer Profile in existing System), which employ classifier-executor-verifier frameworks \citep{Shi2024CHOPSCW} and collaborative generation methods such as Reconcile (a collaborative multi-LLM evaluation framework) and SCRABLE (Self-Correcting Response Approach Based on LLM Evaluation) that use multi-model voting and self-improving loops \citep{10849072, azov2024selfimprovingcustomerreviewresponse}. While these approaches enhance reliability, they remain dependent on large models. A further limitation is the limited availability of publicly accessible customer-service multi-turn QA datasets due to privacy concerns. TelBench introduced the TelTask and TelInstruct datasets \citep{lee-etal-2024-telbench} and TeleEval CS expanded this effort with 90,000 instruction-tuning examples \citep{li2025performance}. However, both benchmarks are restricted to Chinese and Korean and are not explicitly designed for multi-turn customer-service QA. As a result, reproducible evaluation of English multi-turn customer-service QA remains underexplored.

\subsection{Rise of Small Language Models}

The growing interest in SLMs is driven by the limitations of LLMs, including high computational cost, latency and reliance on cloud-based deployment. These challenges are particularly critical in privacy-sensitive and resource-constrained environments. As a result, there is increasing demand for efficient models that can deliver strong performance while remaining cost-effective and deployable on standard hardware. Several SLM families have been introduced in recent years, including SmolLM \citep{allal2025smollm2smolgoesbig}, Qwen \citep{qwen2025qwen25technicalreport, yang2024qwen2technicalreport}, Gemma \citep{gemmateam2025gemma3technicalreport, gemmateam2024gemma2improvingopen}, Phi \citep{microsoft2025phi4mini, abdin2024phi3} and LLaMA \citep{grattafiori2024llama3herdmodels, touvron2023llama2openfoundation}. While differing in size, these models demonstrate strong capabilities in multilingual processing, long-context handling and reasoning with improved efficiency.

Domain-specific adaptation of SLMs has gained increasing attention. In healthcare, models such as BioGPT, PMC-LLaMA, RadPhi2 and CancerGPT have been applied to clinical QA tasks \citep{garg2025risesmalllanguagemodels}. In finance, FinGPT and Instruct-FinGPT have shown strong alignment with domain-specific data \citep{finance}. Customer service, however, remains comparatively underexplored.

Although no specific work has focused on recently introduced instruction-tuned SLMs for multi-turn customer-service QA, some studies have explored customer-service applications using medium-sized SLMs. LoRA-adapted LLaMA-3.1-8B models improved QA accuracy in telecommunications \citep{10971107}, while ChatGLM2-6B achieved high intent accuracy in the electric power sector \citep{11009601}. Studies in banking and restaurant domains reported strong performance using Gemma, Mistral, Falcon and LLaMA-based models \citep{lijaya2025comparative, kdmile}, with real-world prototypes further demonstrating feasibility \citep{10900207}. Nevertheless, systematic evaluation of instruction-tuned SLMs for customer-service tasks involving multi-turn interactions and dialogue continuity remains unexplored.

\subsection{Evaluation Methods Used in Customer Service QA}

Evaluation of customer service QA systems mainly relies on automatic metrics and qualitative assessments. Automatic methods include lexical overlap metrics such as BLEU (Bilingual Evaluation Understudy, a precision-based metric measuring n-gram overlap between generated and reference text) and ROUGE (Recall-Oriented Understudy for Gisting Evaluation, a metric measuring the overlap of words and sequences between generated and reference text) and semantic similarity metrics like BERTScore (a metric that computes token-level semantic similarity between generated and reference text using contextual embeddings from pre-trained encoder-only transformer models) and BARTScore (a metric that uses a pre-trained BART encoder-decoder model to score generated text based on the likelihood of generating the reference text from the candidate response) \citep{BLEU, lin-2004-rouge, zhang2020bertscoreevaluatingtextgeneration, yuan2021bartscoreevaluatinggeneratedtext}. While efficient, these often capture surface-level similarity rather than dialogue coherence. Qualitative evaluations focus on human-centered qualities like correctness, clarity and empathy. Human assessment remains the gold standard, while LLM-as-a-judge frameworks such as G-Eval and PairEval provide scalable alternatives \citep{liu-etal-2023-g, park2024pairevalopendomaindialogueevaluation, gu2025surveyllmasajudge}. However, most studies focus on overall conversation evaluations and have not focused on stage-based evaluations to assess SLMs' abilities across different conversational stages.

Overall, prior work shows that current quantitative and qualitative evaluation approaches do not fully cover all aspects of customer-service QA, particularly in multi-turn settings, due to a lack of benchmarks designed to evaluate dialogue continuity and contextual understanding across conversational turns. These gaps motivate the synthetic dataset construction, fine-tuning pipeline and evaluation framework presented in the following section.

\section{Methodology}
\label{sec:method}

This section outlines the methodological framework adopted to evaluate instruction-tuned SLMs for context-summarized multi-turn customer-service QA. We first describe the
synthetic data construction pipeline designed to address privacy constraints and the lack of publicly
available multi-turn customer-service datasets. We then detail the process of multi-turn conversation construction, context summarization and response refinement used to generate high-quality training
data. Finally, we present the model selection criteria, fine-tuning configuration and inference setup
employed to ensure a controlled and fair comparison between SLMs and LLMs
under identical experimental conditions (code for experiments is available at  footnote\ref{fn:github}).

\footnotetext[1]{\label{fn:github}\url{https://github.com/Lakshan2023/Small_language_models_for_multi_turn_context_summarized_conversations}}

\subsection{Dataset Construction}

\begin{figure*}[!htbp]
    \centering
    \resizebox{\textwidth}{!}{%
        \includegraphics[width=\linewidth, height=4cm, keepaspectratio]{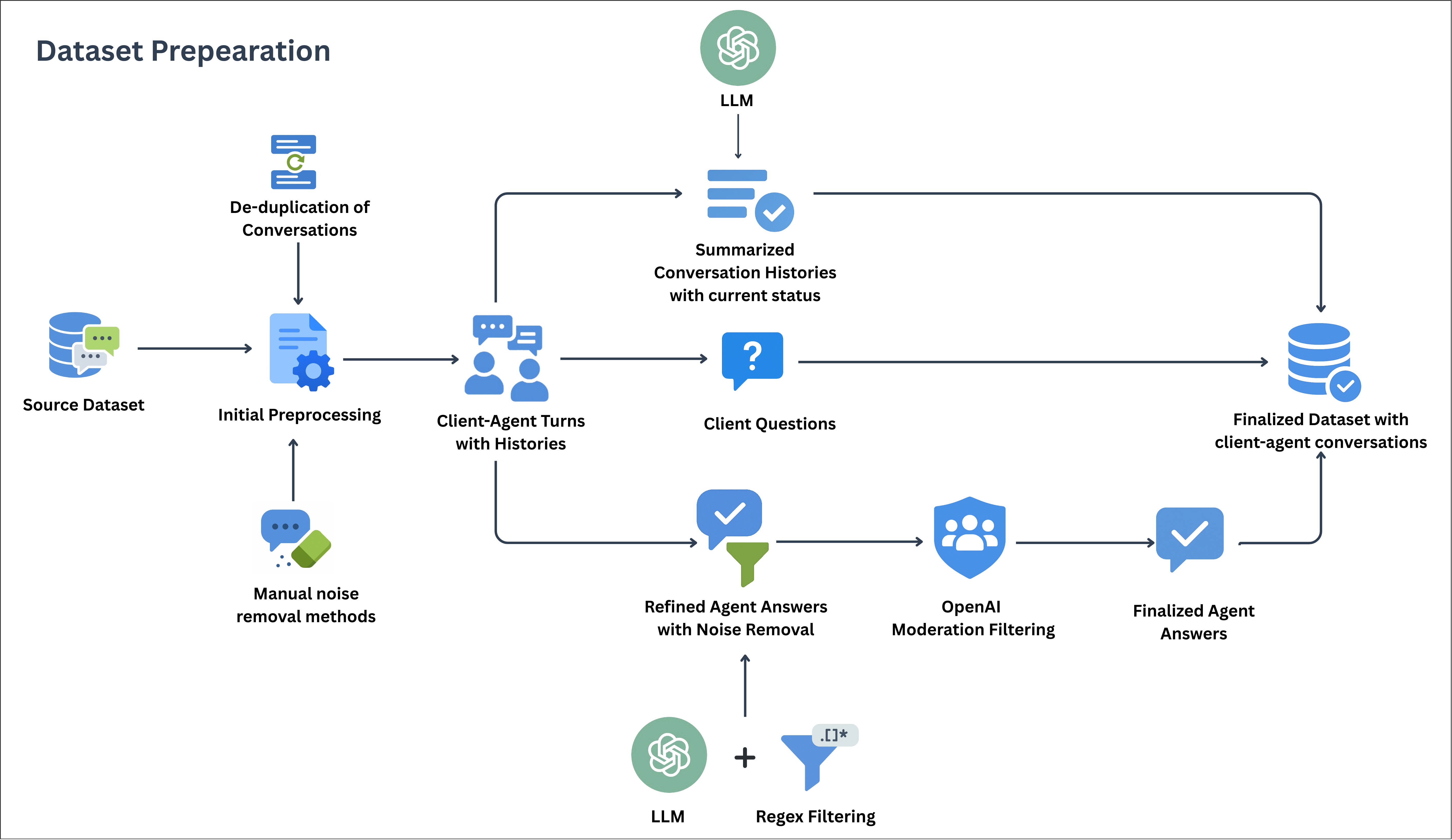}
    }
    \caption{Dataset construction pipeline for context-summarized customer-service QA.}
    \label{fig:pipelines}
\end{figure*}

To address the limited availability of publicly accessible context-summarized multi-turn customer-service data, we constructed a synthetic QA dataset through a controlled pipeline of multi-turn conversation construction, context summarization, response refinement and moderation filtering, designed to preserve essential conversational context while maintaining privacy constraints. The overall dataset construction and processing workflow is illustrated in Figure~\ref{fig:pipelines}.

\subsubsection{\textbf{Initial Data Source}}

We utilized the Customer Service Banking Conversation Corpus from Hugging Face's TalkMap repository as our foundational dataset \citep{talkmap2024banking} (source dataset is available at footnote \ref{fn:talkmap}). While this corpus contained proper conversation sequences, it consisted only of single-turn QA pairs without multi-turn dialogue structure. The initial dataset consisted with 301,822 unique synthetic conversations with 2,880,214 agent messages and 2,651,898 client messages, averaging 18.33 messages per conversation.

\footnotetext[2]{\label{fn:talkmap}\url{https://huggingface.co/datasets/talkmap/banking-conversation-corpus}}

\subsubsection{\textbf{Preprocessing and Filtering}} We applied initial filtering to retain only conversations containing between 5 and 100 turns to ensure realistic conversational depth while excluding extremely short or anomalously long interactions. Very short interactions (less than 5 turns) were excluded as they typically lack sufficient contextual development for evaluating multi-turn reasoning and dialogue continuity. In addition, some conversations in the corpus ended without completing the full interaction flow, which contributed to a number of short conversations falling below this threshold and influenced the selection of this limit. Sample excluded conversations with less than 5 turns are illustrated in Table~\ref{tab:excluded_samples}. Conversations exceeding 100 turns were removed because such extreme lengths are uncommon in real-world customer-service scenarios and are often associated with repetitive exchanges. These long dialogues substantially increase context length, introduce redundancy and reduce the reliability of history summarization and downstream evaluation. The selection of the 5 and 100 turn thresholds was determined through an initial qualitative evaluation of the dataset, followed by consultation with a customer-service domain expert, ensuring that the retained conversations reflect realistic interaction patterns. This filtering resulted in approximately 200,000 conversations used for subsequent processing. Regex-based noise removal was applied to individual conversational turns to eliminate formatting artifacts and non-textual elements.

\begin{table}[!htbp]
\centering
\small
\renewcommand{\arraystretch}{1.5}

\begin{tabular}{p{0.15\textwidth} p{0.8\textwidth}}
\hline

\textbf{Example 01.} &
Agent: "Good morning, thank you for calling Union Financial. My name is John, how may I assist you today?" \newline
Client: "Hi John, I'm calling to inquire about my current savings account balance." \\

\hline

\textbf{Example 02.} &
Agent: "Good Afternoon, you are speaking with Sarah from the Premier Banking support team. How can I help?" \newline
Client: "Good Afternoon Sarah, I would like to check the status of my recent international wire transfer." \newline
Agent: "I can certainly help with that. May I have your full name and account number for verification?" \\

\hline
\end{tabular}

\caption{Sample excluded conversations (less than 5 turns).}
\label{tab:excluded_samples}
\end{table}

\subsubsection{\textbf{Multi-Turn Conversation Construction}} Since the original dataset consisted of isolated single turns that were already in proper sequence, we aggregated all turns belonging to the same conversation to construct multi-turn dialogue instances. De-duplication was subsequently applied to remove redundant conversations, where conversations that appeared more than once in the corpus were discarded to avoid training bias. To create structured training instances, we constructed client-agent pairs with conversational history by randomly partitioning each conversation into early (20\%), middle (70\%) and late (10\%) segments. This conversation splitting strategy ensured balanced coverage of different conversation stages while prioritizing middle turns, which typically contain the most substantive exchanges. As customer-service conversations generally progress with the main interaction occurring in the middle stage, a larger portion of training samples is taken from this stage to improve context understanding during training. The early stage captures the initial problem, while the late stage represents resolution and closing, ensuring learning of the full conversation flow.

\subsubsection{\textbf{Context Summarization}} 
\begin{figure}[!htbp]
    \centering
    \includegraphics[width=\linewidth]{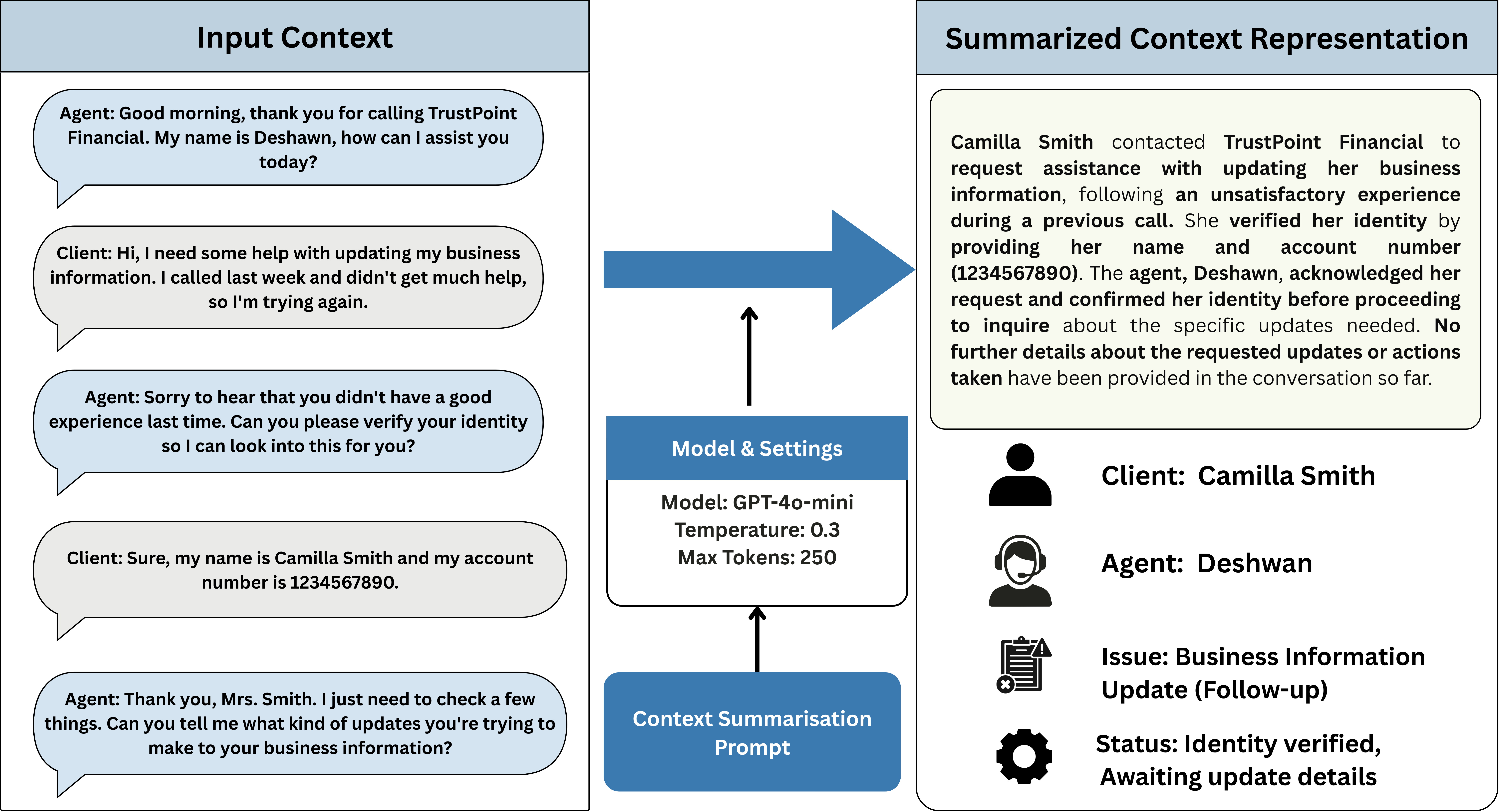}
    \caption{Context summarization process for multi-turn conversation history}
    \label{fig:context_summary_generation}
\end{figure}

SLMs often struggle to maintain context understanding in multi-turn conversational histories. To address this, we apply a history summarization strategy that summarizes prior conversational turns into concise representations while preserving essential information. A specialized prompt instructs the model to generate summaries containing: (1) the client’s primary issue or request and its current status, (2) explicit identification of client and agent names when mentioned, (3) verification steps completed or pending, (4) exact names, account identifiers, dates, amounts and actions taken or agreed upon, (5) commitments, deadlines and scheduled follow-ups and (6) the current conversation status. Context summarization was applied to all conversations using the GPT-4o-mini model with a maximum output length of 250 tokens and a temperature parameter of 0.3. This low temperature was chosen to keep the summaries factually accurate and consistent, since context summarization requires capturing key details from prior dialogue precisely rather than generating creative or varied outputs. The context summarization prompt used to generate the summarized multi-turn conversation histories is provided in Appendix~\ref{app:context_summary_prompt}. In addition, the context summarization process is 
illustrated in Figure~\ref{fig:context_summary_generation}.

\subsubsection{\textbf{Response Refinement}} 
\begin{figure*}[!htbp]
    \centering
    \includegraphics[width=\textwidth]{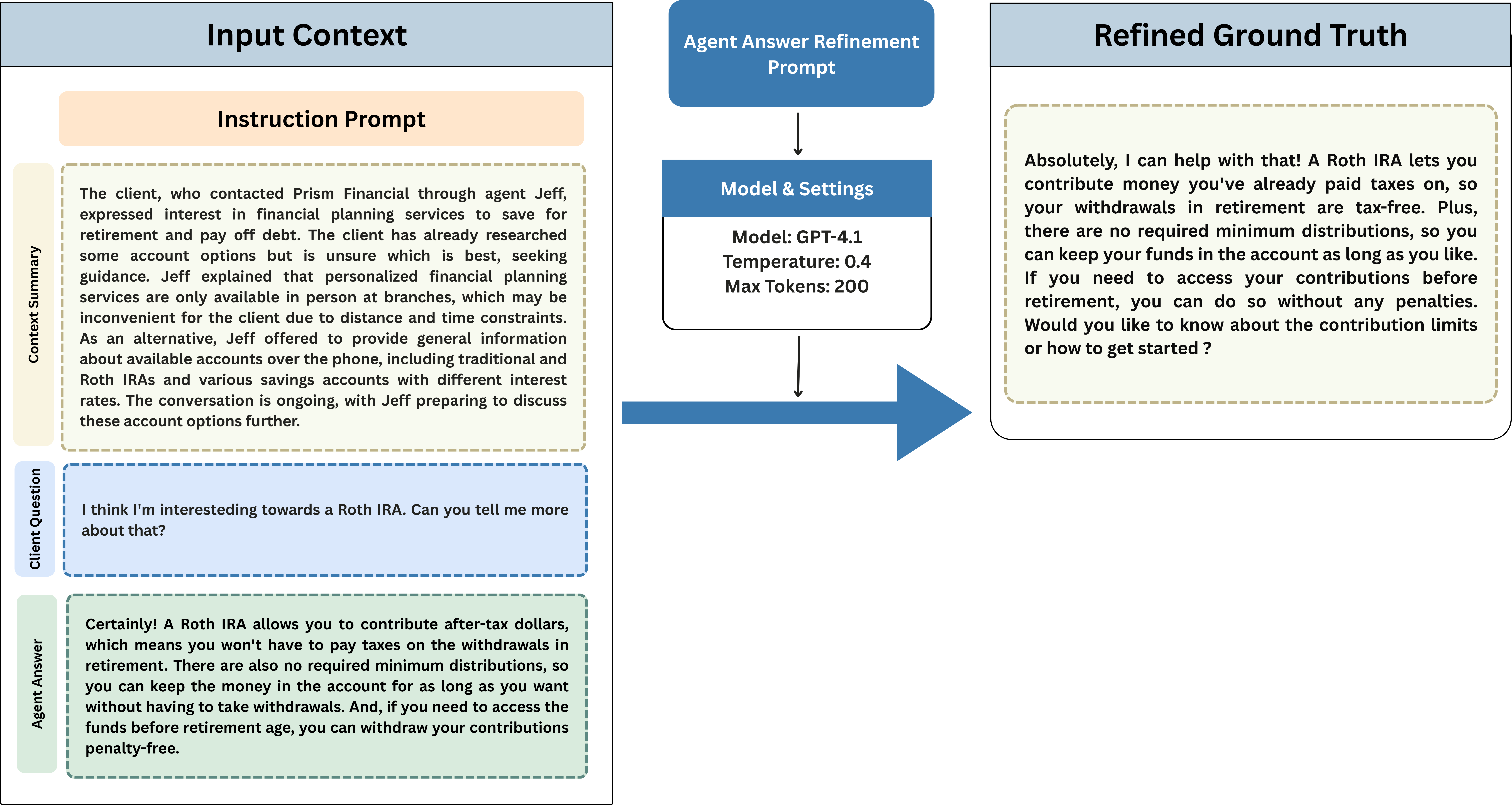}
    \caption{Example instance of GPT-4.1-based response refinement.}
    \label{fig:response_refinement_fig}
\end{figure*}

To improve the qualitative aspects of training data, agent answers in the constructed multi-turn QA dataset were refined using the GPT-4.1 model with a temperature parameter of 0.4. A slightly higher temperature than that used for context summarization was chosen to allow some natural variation in phrasing, while remaining low enough to preserve factual consistency with the original agent answer. The refinement process considered the instruction, client-agent conversation summary, client question and original agent answer. This step improved several qualitative dimensions: naturalness and human-like speaking patterns, appropriate response length according to question complexity, clarity and precision, contextual understanding and coherence with conversational history and removal of noise in the original responses. Since the primary objective of this work is to assess the quality of responses generated by SLMs, ensuring high-quality reference answers in the training data was essential. Following refinement, additional regex-based filtering was applied to remove remaining noise or formatting inconsistencies. The prompt used for response refinement is provided in Appendix~\ref{app:response_refinement}. In addition, Figure~\ref{fig:response_refinement_fig} illustrates an example instance of the response refinement process. Subsequently, we also used OpenAI's Moderation API to flag and filter potentially offensive content from the refined agent answers, ensuring the final dataset adhered to appropriate content standards.

\subsubsection{\textbf{Structured Instance Formation and Dataset Splitting}}

\begin{figure}[!htbp]
    \centering
    \includegraphics[width=12cm, height=8.5cm]{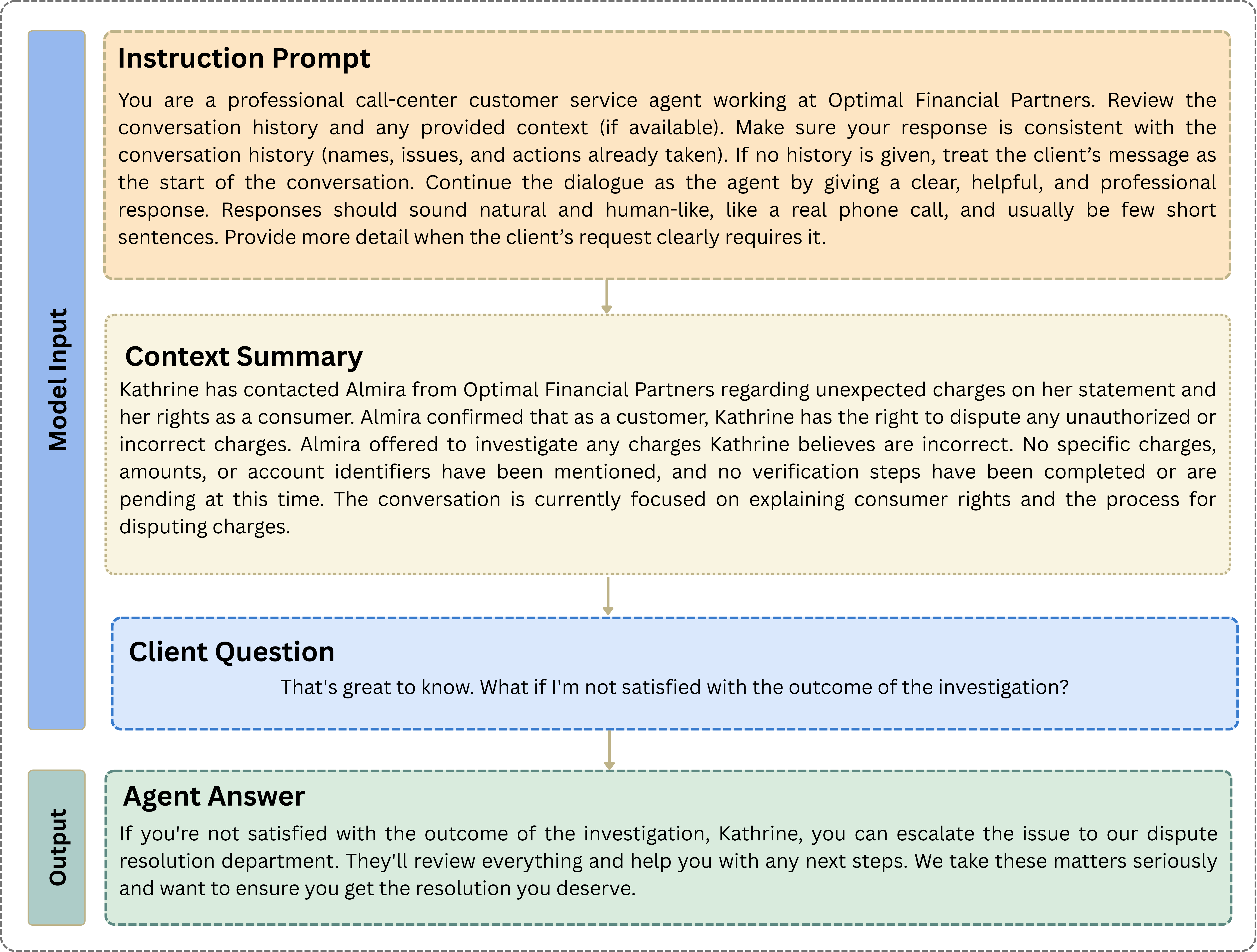}
    \caption{
    Example of a context-summarized multi-turn customer-service QA instance.
    }
    \label{fig:context_summarized_instance}
\end{figure}

Following response refinement, the dataset was organized into structured instances, each comprising summarized history turns, the current client question, the corresponding agent answer and a task-specific instruction prompt. While the prompt structure remained consistent across all instances, the company or financial institution name was varied to maintain diversity, with context summaries and agent answers refined accordingly to remain contextually aligned. An example of a context-summarized multi-turn customer-service QA instance is shown in Figure~\ref{fig:context_summarized_instance}.

The dataset was split into training (70\%), validation (10\%) and test (20\%) sets, with detailed statistics reported in Table~\ref{tab:dataset_stats}. All splits exhibit similar turn-count distributions, ensuring consistent evaluation conditions across experiments. Token counts were computed using GPT-4 tokenization. The constructed synthetic context-summarized multi-turn QA dataset is available at footnote~\ref{fn:hfdataset}.

\footnotetext[3]{\label{fn:hfdataset}\url{https://huggingface.co/datasets/Lakshan2003/customer-support-client-agent-conversations}}

\subsection{Model Selection and Training}

\subsubsection{\textbf{Selected Models}} We evaluate a total of nine fine tuned SLMs spanning multiple parameter ranges. Five models fall within the three to four billion parameter range, namely Qwen 3-4B Instruct, Phi-4 Mini, LLaMA-3.2-3B Instruct, Gemma 3-4B Instruct and SmolLM3-3B. SmolLM3-3B includes enhanced reasoning capabilities; however, explicit reasoning was disabled by not using thinking tags during both training and inference. Gemma 3-4B Instruct is a multimodal model; for this study, fine tuning of the vision components was disabled to focus exclusively on text based conversational understanding. To support comparison across model scales, we additionally include Qwen-3-1.7B Instruct and LLaMA-3.2-1B Instruct from the one to two billion parameter range, as well as Qwen-3-8B Instruct and LLaMA 3.1-8B Instruct from the eight billion parameter range. As Qwen-3-8B Instruct also supports explicit reasoning, reasoning was similarly disabled to ensure consistent evaluation conditions across all fine tuned SLMs.

\subsubsection{\textbf{Training Configuration}} All SLMs included in this study were fine-tuned using Quantized Low-Rank Adaptation (QLoRA) as a parameter-efficient fine-tuning method \citep{dettmers2023QLoRAefficientfinetuningquantized}. QLoRA combines 4-bit quantization with Low-Rank Adaptation, significantly reducing memory requirements while maintaining model performance. Training was conducted using the Unsloth and Hugging Face frameworks.

The models were configured with a maximum sequence length of 512 tokens to accommodate the instruction, summarized conversational history, client question and expected agent response. Prior to adding LoRA adapters, models were quantized to 4-bit precision. The LoRA configuration employed a rank of 16, alpha value of 32 and dropout rate of 0.1, following the recommended configuration provided by Unsloth. A rank of 16 was selected as it provides a good balance between efficiency and model capacity. The alpha was set to twice the rank value, following standard practice for stable training, and a dropout rate of 0.1 was applied to reduce overfitting during fine-tuning. LoRA adapters were applied to all attention and feed-forward projection layers within the Transformer architecture.

Training was performed for 3 epochs using the AdamW 8-bit optimizer with a learning rate of $2 \times 10^{-5}$, weight decay of 0.01 and a warmup ratio of 0.05. A cosine learning rate scheduler was employed to gradually reduce the learning rate over training. All models were trained on an NVIDIA RTX A100 40GB GPU, with training time ranging from 5 to 14 hours per model.

\subsection{Model Inference}

Inference was conducted on the test split containing 36,669 examples. For all models, we set a maximum generation length of 128 tokens to encourage concise responses suitable for customer service interactions.

\subsubsection{\textbf{Small Language Models}} Inference parameters for each SLM were configured according to recommendations provided by the original model publishers to ensure stable and representative performance. For all SLMs, we set the maximum generation length to 128 tokens and enabled sampling during inference. For SmolLM3-3B, we used a temperature of 0.6 with nucleus sampling set to 0.95 and a top-$k$ value of 50. Qwen-3 models, including Qwen-3-4B, Qwen-3-1.7B and Qwen-3-8B, were configured with a temperature of 0.7, nucleus sampling of 0.8, a top-$k$ value of 20 and a minimum probability threshold of 0. Phi-4-Mini employed a temperature of 0.7 with nucleus sampling set to 0.9 and a top-$k$ value of 50. LLaMA-3.2-3B-Instruct, as well as LLaMA-3.2-1B-Instruct and LLaMA-3.1-8B-Instruct, were evaluated using identical decoding parameters, with a temperature of 0.7, nucleus sampling of 0.9 and a top-$k$ value of 50. Gemma-3-4B-Instruct was configured with a temperature of 0.6, nucleus sampling of 0.95, a top-$k$ value of 64 and a repetition penalty of 1.15 to reduce redundant phrasing.

\subsubsection{\textbf{Large Language Models}}

To benchmark the performance of the fine-tuned SLMs, we additionally evaluate three commercial LLMs, namely GPT-4.1, Gemini-2.5-Flash and Virtuoso-Large. All proprietary LLMs were evaluated under identical input prompts and context conditions to ensure fair comparison with the fine-tuned SLMs. GPT-4.1, Virtuoso-Large and Gemini-2.5-Flash were configured with the same decoding parameters (temperature = 0.7, top-$p$ = 0.9). Virtuoso-Large is accessed via the Arcee AI platform and is based on a Qwen-2.5-72B model architecture. As Gemini-2.5-Flash is a reasoning-oriented model, its thinking budget was explicitly set to zero to disable explicit reasoning during inference. For consistency, all LLMs were evaluated using the same maximum generation length of 128 tokens. All LLM inferences were conducted via their respective API endpoints using the same test set and evaluation settings as the fine-tuned SLMs.

\section{Experimental Setup and Evaluation}
\label{sec:experiments}

We evaluate the performance of instruction-tuned SLMs for context-summarized multi-turn customer-service QA using a combination of quantitative and qualitative evaluation methods. The evaluation framework is designed to assess both surface-level alignment with reference answers and higher-level conversational quality, including contextual continuity, tone and task completion. All models are evaluated under identical experimental conditions using the same test set and input format.

\subsection{Quantitative Evaluation}
\label{subsec:quantitative_eval}

Quantitative evaluation is conducted on the full test split of 36,669 examples, focusing on lexical and semantic similarity between generated agent responses and refined reference answers. Although automatic metrics cannot fully capture conversational quality, they provide a reproducible and scalable measure of response alignment.

Lexical similarity is evaluated using ROUGE-L (measures longest common subsequence overlap between generated and reference text) and METEOR (Metric for Evaluation of Translation with Explicit ORdering). ROUGE-L \citep{lin-2004-rouge} measures the longest common subsequence between the generated response and reference answer, capturing structural overlap, while METEOR \citep{banerjee-lavie-2005-meteor} accounts for exact matches, stemming and synonym matches, enabling flexible lexical comparison. Semantic similarity is assessed using BERTScore (F1), BARTScore and cosine similarity between sentence embeddings. BERTScore computes token-level semantic alignment using contextual embeddings \citep{zhang2020bertscoreevaluatingtextgeneration}, BARTScore evaluates the likelihood of generating the reference text from the model output using a pre-trained BART model and cosine similarity captures sentence-level semantic closeness using all-mpnet-base-v2 embeddings. All quantitative metrics are computed using the Hugging Face \texttt{evaluate} library. The official implementation is used for BERTScore, BARTScore is computed with the \texttt{bart-large} model \citep{yuan2021bartscoreevaluatinggeneratedtext} and cosine similarity is calculated over normalized sentence embeddings. Higher values indicate better performance for all metrics except BARTScore, where values closer to zero indicate stronger alignment. Results are reported in Table~\ref{tab:lexical_semantic_results}.

\vspace{5pt}

\begin{table*}[!htbp]
\centering
\small
\setlength{\tabcolsep}{0.5pt}
\renewcommand{\arraystretch}{1.55}
\begin{tabular}{lccccc}
\hline
\textbf{Model} 
& \textbf{ROUGE-L ($\uparrow$)} 
& \textbf{METEOR ($\uparrow$)} 
& \textbf{BARTScore ($\uparrow$)} 
& \textbf{BERTScore F1 ($\uparrow$)} 
& \textbf{Cosine Sim. ($\uparrow$)} \\
\hline
LLaMA-3.2-1B-Instruct 
& 0.2332 
& 0.3032 
& -2.7060 
& 0.8821 
& 0.5909 \\[0.2pt]
Qwen-3-1.7B-Instruct 
& 0.3697 
& 0.4138 
& -2.3096 
& 0.9096 
& 0.6731 \\[0.5pt]
LLaMA-3.2-3B-Instruct 
& 0.3842 
& 0.4471 
& -2.2655 
& 0.9121 
& 0.6958 \\[0.5pt]
SmolLM3-3B-Instruct 
& 0.2393 
& 0.3022 
& -2.7699 
& 0.8830 
& 0.5428 \\[0.5pt]
Phi-4-Mini (3.8B) 
& 0.3747 
& 0.4303 
& -2.2872 
& 0.9107 
& 0.6891 \\[0.5pt]
Qwen-3-4B-Instruct 
& \textbf{0.3959} 
& 0.4455 
& \textbf{-2.2311}
& \textbf{0.9137} 
& 0.6972 \\[0.5pt]
Gemma-3-4B-Instruct 
& 0.2024 
& 0.2782 
& -3.0766 
& 0.8752 
& 0.5134 \\
\hline
LLaMA-3.1-8B-Instruct 
& 0.3940 
& \textbf{0.4569} 
& -2.2332 
& 0.9134 
& \textbf{0.7051} \\[0.5pt]
Qwen-3-8B-Instruct 
& 0.3121 
& 0.3792 
& -2.4970 
& 0.8995 
& 0.6621 \\
\hline
GPT-4.1 
& 0.3038 
& 0.3685 
& -2.5145 
& 0.8994 
& 0.6749 \\[0.5pt]
Gemini-2.5-Flash 
& 0.2771 
& 0.3110 
& -2.6409 
& 0.8942 
& 0.6234 \\[0.5pt]
Virtuoso-Large 
& 0.3161 
& 0.3770 
& -2.4625
& 0.9011 
& 0.6676 \\
\hline
\end{tabular}
\caption{Comparison of lexical and semantic similarity results on the complete test set. Models are grouped by size: small models (<4B), 8B models and commercial large models.}
\label{tab:lexical_semantic_results}
\end{table*}

\vspace{5pt}

\subsection{Conversation Stage Segmentation}

\label{subsec:stage_segmentation}

Before qualitative evaluation, test instances are grouped into three conversation stages: Early, Mid and Late. This stage-based segmentation reflects the natural progression of customer-service interactions, where early-stage turns focus on issue identification, mid-stage turns contain the core interaction and information exchange and late-stage turns emphasize resolution and closure. Conversation stage assignment is performed using the GPT-4.1-mini model with a temperature of 0 to ensure deterministic and reproducible outputs. This segmentation enables controlled sampling, balanced stage-wise coverage and targeted analysis of model behavior under varying contextual demands. For both human and LLM-as-a-judge evaluation, samples are selected using a fixed ratio of 10\% early-stage, 80\% mid-stage and 10\% late-stage instances, placing greater emphasis on mid-stage interactions that require stronger contextual reasoning and dialogue continuity. Based on this segmentation, we conduct a stage-based qualitative analysis that evaluates model performance at the Early, Mid and Late stages of customer-service conversations and apply this analysis consistently across both evaluation settings.

\vspace{-5pt}

\subsection{LLM-as-a-Judge Evaluation}
\label{subsec:llm_judge_eval}
To assess conversational quality at scale, we employ an LLM-as-a-judge evaluation framework based on the G-Eval methodology ~\citep{liu-etal-2023-g}. A specialized prompt is used to score generated responses across four qualitative dimensions: Human-Likeness, Continuity and Context Understanding, Tone and Clarity and Task Appropriateness. Each dimension is scored independently on a 1-5 Likert scale, where higher scores indicate better performance. Claude Sonnet 4.5 was selected as the judge model because it was not among the models being evaluated, which helps avoid any bias a model might have toward its own outputs. A temperature of 0 was used to minimize output variability, ensuring that the judge produces consistent and reproducible scores and judgments across all evaluated models. For each response, the judge is provided with the summarized conversation history, the client question, a reference agent response supplied only as guidance and the model-generated response. LLM-as-a-judge evaluation is performed on 6,000 randomly sampled test instances per model. Scores are averaged across all evaluated instances to obtain overall performance metrics, which are reported in Table~\ref{tab:llm_judge_eval}. 

\begin{table*}[!htbp]
\centering
\small
\setlength{\tabcolsep}{5pt}
\renewcommand{\arraystretch}{1.45}
\begin{tabular}{lccccc}
\hline
\textbf{Model} 
& \textbf{Human} 
& \textbf{Continuity} 
& \textbf{Tone} 
& \textbf{Task} 
& \textbf{Overall} \\
& \textbf{Likeness} 
& \textbf{\& Context Understanding} 
& \textbf{\& Clarity} 
& \textbf{Appropriateness} 
& \textbf{Mean} \\
\hline
LLaMA-3.2-1B-Instruct 
& 3.165 
& 2.358 
& 3.342 
& 2.171 
& 2.759 \\[0.5pt]
Qwen-3-1.7B-Instruct 
& 3.738 
& 3.362 
& 3.818 
& 2.994 
& 3.478 \\[0.5pt]
SmolLM3-3B-Instruct 
& 2.654 
& 1.772 
& 2.717 
& 1.696 
& 2.210 \\[0.5pt]
LLaMA-3.2-3B-Instruct 
& 4.075 
& 3.480 
& 4.105 
& 3.212 
& 3.718 \\[0.5pt]
Phi-4-Mini (3.8 B)
& 3.988 
& 3.360 
& 4.034 
& 3.093 
& 3.619 \\[0.5pt]
Qwen-3-4B-Instruct 
& 4.044 
& 3.430 
& 4.071 
& 3.170 
& 3.679 \\[0.5pt]
Gemma-3-4B-Instruct 
& 2.582 
& 1.729 
& 2.597 
& 1.673 
& 2.145 \\
\hline
LLaMA-3.1-8B-Instruct 
& 4.115 
& 3.591 
& 4.149 
& 3.322 
& 3.794 \\[0.5pt]
Qwen-3-8B-Instruct 
& 3.950 
& 3.648 
& 4.067 
& 3.306 
& 3.743 \\
\hline
GPT-4.1 
& \textbf{4.316} 
& \textbf{4.079} 
& \textbf{4.381} 
& \textbf{3.808} 
& \textbf{4.146} \\[0.3pt]
Gemini-2.5-Flash 
& 4.054 
& 3.742 
& 4.101 
& 3.180 
& 3.769 \\[0.5pt]
Virtuoso-Large 
& 4.171 
& 3.864 
& 4.204 
& 3.530 
& 3.942 \\
\hline
\end{tabular}
\caption{Overall LLM-as-a-judge evaluation results across four qualitative dimensions using a 5-point Likert scale. Models are grouped by size: small models (<4B), 8B models and commercial LLMs.}
\label{tab:llm_judge_eval}
\end{table*}

\subsection{Human Evaluation}
\label{subsec:human_eval}

Human evaluation was conducted to provide a gold-standard assessment of conversational quality. Due to limited evaluation resources, this analysis was restricted to SLMs in the 3-4B parameter range and the selected commercial LLMs. Three human evaluators independently assessed model-generated responses using the same qualitative dimensions as the LLM-as-a-judge evaluation. Evaluators were provided with the summarized conversation history, the client question and the model-generated response and model identities were hidden to avoid bias. Scores were assigned on a 1-5 Likert scale and averaged across evaluators. This setup enabled both overall analysis of conversational performance and allowed direct comparison with the LLM-as-a-judge results. Aggregated overall human evaluation results are reported in Table~\ref{tab:human_eval}.

\begin{table*}[!htbp]
\centering
\small
\setlength{\tabcolsep}{5pt}
\renewcommand{\arraystretch}{1.65}
\begin{tabular}{lccccc}
\hline
\textbf{Model} 
& \textbf{Human} 
& \textbf{Continuity} 
& \textbf{Tone} 
& \textbf{Task} 
& \textbf{Overall} \\
& \textbf{Likeness} 
& \textbf{\& Context Understanding} 
& \textbf{\& Clarity} 
& \textbf{Appropriateness} 
& \textbf{Mean} \\
\hline
SmolLM3-3B-Instruct 
& 3.003 
& 2.615 
& 2.965 
& 2.261 
& 2.711 \\[0.5pt]
LLaMA-3.2-3B-Instruct 
& 4.250 
& 4.325 
& 4.286 
& 3.721 
& 4.146 \\[0.5pt]
Phi-4-Mini (3.8B)
& 4.164 
& 4.303 
& 4.215 
& 3.553 
& 4.059 \\[0.5pt]
Qwen-3-4B-Instruct 
& 4.203 
& 4.264 
& 4.230 
& 3.579 
& 4.069 \\[0.5pt]
Gemma-3-4B-Instruct 
& 3.110 
& 2.520 
& 2.968 
& 2.146 
& 2.686 \\
\hline
GPT-4.1 
& \textbf{4.674} 
& \textbf{4.827} 
& \textbf{4.722} 
& \textbf{4.286} 
& \textbf{4.627} \\[0.5pt]
Gemini-2.5-Flash 
& 4.181 
& 4.567 
& 4.247 
& 3.770 
& 4.191 \\[0.5pt]
Virtuoso-Large 
& 4.507 
& 4.726 
& 4.637 
& 4.249 
& 4.529 \\
\hline
\end{tabular}
\caption{Overall Human evaluation results across four qualitative dimensions using a 5-point Likert scale. Models are grouped by size: small models (<4B) and commercial LLMs.}
\label{tab:human_eval}
\end{table*}

\subsection{Pairwise Evaluation}
\label{subsec:pairwise_eval}

Pairwise evaluation is conducted between selected high-performing SLMs and commercial LLMs. For each input, two responses (A and B) are compared directly and the judge selects the better response overall \citep{park2024pairevalopendomaindialogueevaluation, liu2025aligninghumanjudgementrole}. The evaluation uses 1,000 test instances with responses generated by the selected models and Claude Haiku~4.5 as the judge, with the temperature set to 0. To mitigate positional bias, each instance is evaluated twice by swapping the A and B ordering. If the same model is preferred in both orderings, the outcome is recorded as a win for that model. If the preferred model differs across orderings, the instance is treated as a tie. The judge applies the same four qualitative criteria used in the Likert-scale evaluations and outputs a single winner per comparison. This evaluation provides a direct preference-based comparison between LLMs and SLMs that complements score-based assessments by highlighting relative response quality under identical prompts. Results are reported as win and tie percentages for each model pair and summarized in Table~\ref{tab:pairwise_eval}. 

\begin{table*}[!htb]
\centering
\small
\setlength{\tabcolsep}{7pt}
\renewcommand{\arraystretch}{1.65}
\begin{tabular}{llccc}
\hline
\textbf{LLM} 
& \textbf{SLM} 
& \textbf{LLM Wins (\%)} 
& \textbf{SLM Wins (\%)} 
& \textbf{Ties (\%)} \\
\hline
Gemini-2.5-Flash 
& LLaMA-3.2-1B-Instruct 
& 43.10 & 38.60 & 18.30 \\
Gemini-2.5-Flash 
& Qwen-3-1.7B-Instruct 
& 49.80 & 33.50 & 16.70 \\
Gemini-2.5-Flash 
& LLaMA-3.2-3B-Instruct 
& 37.40 & 49.70 & 12.90 \\
Gemini-2.5-Flash 
& Phi-4-Mini (3.8 B) 
& 40.50 & 43.90 & 15.60 \\
Gemini-2.5-Flash 
& Qwen-3-4B-Instruct 
& 41.20 & 45.00 & 13.80 \\
Gemini-2.5-Flash 
& LLaMA-3.1-8B-Instruct 
& 28.60 & 52.90 & 18.50 \\
Gemini-2.5-Flash 
& Qwen-3-8B-Instruct 
& 26.60 & 55.80 & 17.60 \\
\hline
GPT-4.1 
& LLaMA-3.2-1B-Instruct 
& 68.60 & 15.70 & 15.70 \\
GPT-4.1 
& Qwen-3-1.7B-Instruct 
& 79.00 & 8.70 & 12.30 \\
GPT-4.1 
& LLaMA-3.2-3B-Instruct 
& 67.00 & 17.90 & 15.10 \\
GPT-4.1 
& Phi-4-Mini (3.8 B) 
& 72.50 & 14.90 & 12.60 \\
GPT-4.1 
& Qwen-3-4B-Instruct 
& 70.90 & 14.60 & 14.50 \\
GPT-4.1 
& LLaMA-3.1-8B-Instruct 
& 61.30 & 19.00 & 19.70 \\
GPT-4.1 
& Qwen-3-8B-Instruct 
& 54.20 & 23.80 & 22.00 \\
\hline
Virtuoso-Large 
& LLaMA-3.2-1B-Instruct 
& 61.60 & 19.80 & 18.60 \\
Virtuoso-Large 
& Qwen-3-1.7B-Instruct 
& 73.40 & 12.80 & 13.80 \\
Virtuoso-Large 
& LLaMA-3.2-3B-Instruct 
& 56.90 & 24.60 & 18.50 \\
Virtuoso-Large 
& Phi-4-Mini (3.8 B) 
& 64.70 & 19.90 & 15.40 \\
Virtuoso-Large 
& Qwen-3-4B-Instruct 
& 62.60 & 21.80 & 15.60 \\
Virtuoso-Large 
& LLaMA-3.1-8B-Instruct 
& 54.70 & 28.00 & 17.30 \\
Virtuoso-Large 
& Qwen-3-8B-Instruct 
& 46.70 & 31.90 & 21.40 \\
\hline
\end{tabular}
\caption{Pairwise LLM vs. SLM evaluation results expressed as win percentages.}
\label{tab:pairwise_eval}
\end{table*}

\subsection{Conversational Stage-based Evaluation}
\label{stage_based}

Most existing customer-service QA research reports qualitative results at the overall conversation level. However, the performance of fine-tuned models can vary across different conversation stages (Early, Mid and Late). To capture these variations, we apply the same qualitative evaluation approach across stage-wise grouped conversations, following the segmentation procedure described in Section~\ref{subsec:stage_segmentation}.

\subsubsection{\textbf{Conversational Stage-based LLM-as-a-Judge Evaluation}}

The stage-wise LLM-as-a-judge evaluation is conducted using Claude Sonnet 4.5 as the judge on the same set of 6,000 responses used in the overall LLM-as-a-judge evaluation, segmented into 600 Early-stage, 4,800 Mid-stage and 600 Late-stage responses, following the segmentation ratio described in Section~\ref{subsec:stage_segmentation}. Scores are assigned independently across four qualitative dimensions using a 
1-5 Likert scale, following the same criteria described in 
Section~\ref{subsec:llm_judge_eval}. Table ~\ref{tab:llm_judge_stagewise} reports the LLM-as-a-judge-based stage-wise results separately for each conversation phase, enabling a direct comparison of model behavior across Early, Mid, and Late interactions. This stage-based analysis enables identification of models that perform 
consistently across stages and models that show performance variations 
across different conversation phases.

\begin{table*}[!htbp]
\centering
\scriptsize  % Smaller font
\setlength{\tabcolsep}{2pt}  % Tighter column spacing
\renewcommand{\arraystretch}{1.45}  % Reduced row height
\begin{tabular}{c l c c c c c}
\hline
\textbf{Conversation} & \textbf{Model} & \textbf{Human} & \textbf{Continuity} & \textbf{Tone} & \textbf{Task} & \textbf{Overall} \\
\textbf{Stage} & & \textbf{Likeness} & \textbf{\& Context Understanding} & \textbf{\& Clarity} & \textbf{Appropriateness} & \textbf{Mean} \\
\hline
\multirow{12}{*}{\rotatebox{90}{\textbf{Early-stage}}} 
& LLaMA-3.2-1B-Instruct & 3.625 & 2.927 & 3.725 & 2.583 & 3.215 \\
& Qwen-3-1.7B-Instruct & 3.880 & 3.508 & 3.957 & 3.145 & 3.622 \\
& LLaMA-3.2-3B-Instruct & 4.125 & 3.625 & 4.160 & 3.288 & 3.800 \\
& SmolLM3-3B & 2.227 & 1.578 & 2.290 & 1.503 & 1.900 \\
& Qwen-3-4B-Instruct & 4.100 & 3.575 & 4.130 & 3.253 & 3.764 \\
& Phi-4-Mini & 4.058 & 3.473 & 4.113 & 3.158 & 3.700 \\
& Gemma-3-4B-Instruct & 2.517 & 1.553 & 2.557 & 1.508 & 2.034 \\
\cline{2-7}
& LLaMA-3.1-8B-Instruct & 4.152 & 3.613 & 4.202 & 3.310 & 3.819 \\
& Qwen-3-8B-Instruct & 3.932 & 3.513 & 4.038 & 3.175 & 3.665 \\
\cline{2-7}
& GPT-4.1 & \textbf{4.310} & \textbf{4.018} & \textbf{4.383} & \textbf{3.715} & \textbf{4.106} \\
& Virtuoso-Large & 4.157 & 3.750 & 4.185 & 3.425 & 3.879 \\
& Gemini-2.5-Flash & 4.143 & 3.768 & 4.185 & 3.350 & 3.862 \\
\hline
\multirow{12}{*}{\rotatebox{90}{\textbf{Mid-Stage}}} 
& LLaMA-3.2-1B-Instruct & 3.139 & 2.327 & 3.315 & 2.150 & 2.733 \\
& Qwen-3-1.7B-Instruct & 3.684 & 3.277 & 3.766 & 2.900 & 3.407 \\
& LLaMA-3.2-3B-Instruct & 4.034 & 3.388 & 4.065 & 3.104 & 3.648 \\
& SmolLM3-3B & 2.738 & 1.806 & 2.798 & 1.719 & 2.265 \\
& Qwen-3-4B-Instruct & 4.001 & 3.342 & 4.030 & 3.061 & 3.609 \\
& Phi-4-Mini & 3.943 & 3.267 & 3.991 & 2.980 & 3.545 \\
& Gemma-3-4B-Instruct & 2.618 & 1.737 & 2.624 & 1.659 & 2.160 \\
\cline{2-7}
& LLaMA-3.1-8B-Instruct & 4.077 & 3.525 & 4.111 & 3.236 & 3.737 \\
& Qwen-3-8B-Instruct & 3.924 & 3.625 & 4.052 & 3.249 & 3.712 \\
\cline{2-7}
& GPT-4.1 & \textbf{4.286} & \textbf{4.056} & \textbf{4.355} & \textbf{3.764} & \textbf{4.115} \\
& Virtuoso-Large & 4.137 & 3.835 & 4.173 & 3.470 & 3.904 \\
& Gemini-2.5-Flash & 4.039 & 3.759 & 4.081 & 3.127 & 3.752 \\
\hline
\multirow{12}{*}{\rotatebox{90}{\textbf{Late-stage}}} 
& LLaMA-3.2-1B-Instruct & 2.910 & 2.033 & 3.173 & 1.930 & 2.512 \\
& Qwen-3-1.7B-Instruct & 4.032 & 3.902 & 4.095 & 3.591 & 3.905 \\
& LLaMA-3.2-3B-Instruct & 4.357 & 4.068 & 4.373 & 4.002 & 4.200 \\
& SmolLM3-3B & 2.410 & 1.688 & 2.500 & 1.700 & 2.074 \\
& Qwen-3-4B-Instruct & 4.335 & 3.993 & 4.340 & 3.950 & 4.154 \\
& Phi-4-Mini & 4.273 & 3.983 & 4.297 & 3.932 & 4.121 \\
& Gemma-3-4B-Instruct & 2.363 & 1.843 & 2.418 & 1.945 & 2.142 \\
\cline{2-7}
& LLaMA-3.1-8B-Instruct & 4.386 & 4.102 & 4.397 & 4.030 & 4.229 \\
& Qwen-3-8B-Instruct & 4.182 & 3.967 & 4.212 & 3.885 & 4.062 \\
\cline{2-7}
& GPT-4.1 & \textbf{4.563} & \textbf{4.323} & \textbf{4.588} & \textbf{4.253} & \textbf{4.432} \\
& Virtuoso-Large & 4.453 & 4.212 & 4.468 & 4.115 & 4.312 \\
& Gemini-2.5-Flash & 4.085 & 3.577 & 4.177 & 3.433 & 3.818 \\
\hline
\end{tabular}
\caption{Stage-wise LLM-as-a-judge evaluation results across early, mid and late-stage customer-service interactions using a 5-point Likert scale. Scores are averaged over 6,000 evaluation samples, with 600 early-stage, 4,800 mid-stage and 600 late-stage instances.}
\label{tab:llm_judge_stagewise}
\end{table*}

\subsubsection{\textbf{Conversational Stage-based Human Evaluation}}

The stage-wise human evaluation is conducted on the same set of 500 responses per model used in the overall human evaluation,
segmented into 50 Early-stage, 400 Mid-stage and 50 Late-stage instances, following the same assessment
criteria and scoring procedure described in Section~\ref{subsec:human_eval}. However, due to limitations in human evaluation resources, this experiment is also conducted only on commercial LLMs and SLMs in the 3-4B parameter range. Scores are reported separately for each conversation
stage and qualitative dimensions. Stage-wise human evaluation results are reported in Table~\ref{tab:human_eval_stagewise}, enabling
direct cross-validation of stage-wise patterns observed in the LLM-as-a-judge evaluation and providing a
gold-standard assessment of model behavior across different conversation phases.

\begin{table*}[!ht]
\centering
\small
\setlength{\tabcolsep}{2pt}
\renewcommand{\arraystretch}{1.45}
\begin{tabular}{c l c c c c c}
\hline
\textbf{Stage} & \textbf{Model} &
\textbf{Human} &
\textbf{Continuity} &
\textbf{Tone} &
\textbf{Task} &
\textbf{Overall} \\
& &
\textbf{Likeness} &
\textbf{\& Context} &
\textbf{\& Clarity} &
\textbf{Appropriateness} &
\textbf{Mean} \\
\hline
\multirow{8}{*}{\textbf{Early-Stage}}
& LLaMA-3.2-3B-Instruct & 4.107 & 4.227 & 4.093 & 3.593 & 4.005 \\
& SmolLM3-3B & 2.513 & 2.173 & 2.487 & 1.833 & 2.252 \\
& Qwen-3-4B-Instruct & 4.073 & 4.060 & 4.093 & 3.433 & 3.915 \\
& Phi-4-Mini & 3.940 & 4.147 & 4.000 & 3.440 & 3.882 \\
& Gemma-3-4B-Instruct & 3.013 & 2.360 & 2.887 & 1.953 & 2.553 \\
\cline{2-7}
& GPT-4.1 & \textbf{4.467} & \textbf{4.753} & \textbf{4.540} & \textbf{4.073} & \textbf{4.458} \\
& Virtuoso-Large & 4.353 & 4.680 & 4.507 & 4.100 & 4.410 \\
& Gemini-2.5-Flash & 4.187 & 4.527 & 4.240 & 3.933 & 4.222 \\
\hline
\multirow{8}{*}{\textbf{Mid-stage}}
& LLaMA-3.2-3B-Instruct & 4.217 & 4.291 & 4.258 & 3.650 & 4.104 \\
& SmolLM3-3B & 3.098 & 2.694 & 3.059 & 2.323 & 2.793 \\
& Qwen-3-4B-Instruct & 4.164 & 4.249 & 4.197 & 3.510 & 4.030 \\
& Phi-4-Mini & 4.138 & 4.274 & 4.187 & 3.478 & 4.019 \\
& Gemma-3-4B-Instruct & 3.141 & 2.531 & 3.002 & 2.143 & 2.704 \\
\cline{2-7}
& GPT-4.1 & \textbf{4.675} & \textbf{4.846} & \textbf{4.725} & \textbf{4.283} & \textbf{4.632} \\
& Virtuoso-Large & 4.489 & 4.723 & 4.631 & 4.223 & 4.517 \\
& Gemini-2.5-Flash & 4.159 & 4.583 & 4.233 & 3.731 & 4.177 \\
\hline
\multirow{8}{*}{\textbf{Late-stage}}
& LLaMA-3.2-3B-Instruct & 4.660 & 4.700 & 4.700 & 4.420 & 4.620 \\
& SmolLM3-3B & 2.733 & 2.427 & 2.693 & 2.193 & 2.512 \\
& Qwen-3-4B-Instruct & 4.640 & 4.587 & 4.627 & 4.273 & 4.532 \\
& Phi-4-Mini & 4.593 & 4.687 & 4.653 & 4.260 & 4.548 \\
& Gemma-3-4B-Instruct & 2.960 & 2.593 & 2.780 & 2.360 & 2.673 \\
\cline{2-7}
& GPT-4.1 & \textbf{4.867} & \textbf{4.753} & \textbf{4.880} & \textbf{4.520} & \textbf{4.755} \\
& Virtuoso-Large & 4.800 & 4.793 & 4.820 & 4.600 & 4.753 \\
& Gemini-2.5-Flash & 4.347 & 4.480 & 4.367 & 3.920 & 4.278 \\
\hline
\end{tabular}
\caption{Stage-wise human evaluation results across early, mid and late-stage customer-service interactions using a 5-point Likert scale. Scores are averaged over 500 evaluation samples per model, consisting of 50 early-stage, 400 mid-stage and 50 late-stage instances.}
\label{tab:human_eval_stagewise}
\end{table*}

\subsubsection{\textbf{Conversational Stage-based Pairwise Evaluation}}

The stage-wise pairwise evaluation is conducted on the same 1,000 randomly selected test instances used in Section~\ref{subsec:pairwise_eval}, with responses evaluated using Claude Haiku 4.5 as the judge (temperature set to 0). These instances are segmented into Early, Mid and Late stages following the procedure described in Section~\ref{subsec:stage_segmentation}, with a distribution of 80.4\% Mid-stage, 9.9\% Early-stage and 9.7\% Late-stage samples. Win and tie percentages are computed separately for each conversation stage to directly compare SLMs with commercial LLMs. Table~\ref{tab:pairwise_conversation_stages} reports the stage-wise results, enabling comparison of model performance across different conversation phases.

\begin{table*}[!htbp]
\centering
\small % Slightly larger than footnotesize
\setlength{\tabcolsep}{8pt}  % Increased column spacing
\renewcommand{\arraystretch}{1.5}  % Increased row spacing
\begin{tabular}{l|ccc|ccc|ccc}
\hline
\multirow{2}{*}{\textbf{Tested SLM}} & \multicolumn{3}{c|}{\textbf{Early-Stage}} & \multicolumn{3}{c|}{\textbf{Mid-Stage}} & \multicolumn{3}{c}{\textbf{Late-Stage}} \\
\cline{2-10}
& \textbf{LLM} & \textbf{SLM} & \textbf{Tie} & \textbf{LLM} & \textbf{SLM} & \textbf{Tie} & \textbf{LLM} & \textbf{SLM} & \textbf{Tie} \\
\hline
\multicolumn{10}{c}{\textit{GPT-4.1}} \\
\hline
LLaMA-3.2-1B-Instruct & 65.66 & 17.17 & 17.17 & 72.01 & 12.94 & 15.05 & 43.30 & 37.11 & 19.59 \\
Qwen-3-1.7B-Instruct & 75.76 & 11.11 & 13.13 & 82.34 & 6.84 & 10.82 & 54.64 & 21.65 & 23.71 \\
LLaMA-3.2-3B-Instruct & 72.73 & 18.18 & 9.09 & 67.91 & 16.54 & 15.55 & 53.61 & 28.87 & 17.53 \\
Phi-4-Mini (3.8 B) & 71.72 & 13.13 & 15.15 & 74.75 & 13.68 & 11.57 & 54.64 & 26.80 & 18.56 \\
Qwen-3-4B-Instruct & 66.67 & 14.14 & 19.19 & 72.89 & 13.43 & 13.68 & 58.76 & 24.74 & 16.49 \\
LLaMA-3.1-8B-Instruct & 54.55 & 22.22 & 23.23 & 64.05 & 18.16 & 17.79 & 45.36 & 22.68 & 31.96 \\
Qwen-3-8B-Instruct & 54.55 & 19.19 & 26.26 & 54.35 & 24.25 & 21.39 & 52.58 & 24.74 & 22.68 \\
\hline
\multicolumn{10}{c}{\textit{Virtuoso-Large}} \\
\hline
LLaMA-3.2-1B-Instruct & 57.58 & 25.25 & 17.17 & 64.43 & 17.29 & 18.28 & 42.27 & 35.05 & 22.68 \\
Qwen-3-1.7B-Instruct & 65.66 & 14.14 & 20.20 & 76.37 & 12.81 & 10.82 & 56.70 & 11.34 & 31.96 \\
LLaMA-3.2-3B-Instruct & 58.59 & 27.27 & 14.14 & 57.84 & 23.63 & 18.53 & 47.42 & 29.90 & 22.68 \\
Phi-4-Mini (3.8 B) & 57.58 & 26.26 & 16.16 & 67.79 & 18.91 & 13.31 & 46.39 & 21.65 & 31.96 \\
Qwen-3-4B-Instruct & 56.57 & 27.27 & 16.16 & 65.17 & 20.52 & 14.30 & 47.42 & 26.80 & 25.77 \\
LLaMA-3.1-8B-Instruct & 47.47 & 34.34 & 18.18 & 57.21 & 26.12 & 16.67 & 41.24 & 37.11 & 21.65 \\
Qwen-3-8B-Instruct & 49.49 & 31.31 & 19.19 & 45.27 & 33.33 & 21.39 & 55.67 & 20.62 & 23.71 \\
\hline
\multicolumn{10}{c}{\textit{Gemini-2.5-Flash}} \\
\hline
LLaMA-3.2-1B-Instruct & 49.49 & 35.35 & 15.15 & 43.91 & 38.68 & 17.41 & 29.90 & 41.24 & 28.87 \\
Qwen-3-1.7B-Instruct & 56.57 & 28.28 & 15.15 & 49.50 & 34.08 & 16.42 & 45.36 & 34.02 & 20.62 \\
LLaMA-3.2-3B-Instruct & 42.42 & 44.44 & 13.13 & 36.82 & 50.62 & 12.56 & 37.11 & 47.42 & 15.46 \\
Phi-4-Mini (3.8 B) & 45.45 & 37.37 & 17.17 & 40.67 & 43.66 & 15.67 & 34.02 & 52.58 & 13.40 \\
Qwen-3-4B-Instruct & 49.49 & 36.36 & 14.14 & 40.67 & 46.02 & 13.31 & 37.11 & 45.36 & 17.53 \\
LLaMA-3.1-8B-Instruct & 31.31 & 40.40 & 28.28 & 26.99 & 55.97 & 17.04 & 39.18 & 40.21 & 20.62 \\
Qwen-3-8B-Instruct & 44.44 & 38.38 & 17.17 & 23.13 & 60.45 & 16.42 & 37.11 & 35.05 & 27.84 \\
\hline
\end{tabular}
\caption{Win and tie percentages for pairwise comparisons between selected high-performing SLMs and commercial LLMs across Early, Mid and Late conversation stages, using Claude Haiku 4.5 as the judge. Results are computed on 1,000 randomly selected test instances, distributed as 80.4\% mid-stage, 9.9\% early-stage and 9.7\% late-stage samples from the test corpus.}
\label{tab:pairwise_conversation_stages}
\end{table*}

\section{Discussion on Obtained Results}
\label{subsec:obtained_results}

The evaluation framework applied in this study provides comprehensive insight into the performance of fine-tuned instruction-tuned SLMs for context-summarized multi-turn customer-service QA. Although the majority of fine-tuned SLMs obtain higher scores than commercial LLMs on quantitative metrics (Table~\ref{tab:lexical_semantic_results}), this primarily reflects closer alignment with the reference response distribution used for evaluation rather than overall conversational quality. Commercial LLMs achieve higher scores in LLM-as-a-judge (Table~\ref{tab:llm_judge_eval}) and human evaluation (Table~\ref{tab:human_eval}), demonstrating that lexical and semantic similarity 
metrics alone are insufficient for evaluating context-summarized multi-turn customer-service QA.

Quantitative evaluation shows that the strongest fine-tuned SLMs achieve consistently competitive performance across lexical and semantic metrics (Table~\ref{tab:lexical_semantic_results}). Qwen-3-4B-Instruct attains the highest scores in ROUGE-L (0.3959), BARTScore (-2.2311) and BERTScore F1 (0.9137), while LLaMA-3.1-8B-Instruct records the highest METEOR score (0.4569) and cosine similarity (0.7051). LLaMA-3.2-3B-Instruct (ROUGE-L: 0.3842, BERTScore F1: 0.9121) and Phi-4-Mini (ROUGE-L: 0.3747, BERTScore F1: 0.9107) also perform competitively, indicating strong lexical overlap and semantic alignment. Within the less than 2B parameter range, Qwen-3-1.7B-Instruct (ROUGE-L: 0.3697, BERTScore F1: 0.9096) outperforms LLaMA-3.2-1B-Instruct (ROUGE-L: 0.2332, BERTScore F1: 0.8821). In addition, SmolLM3-3B-Instruct (ROUGE-L: 0.2393, cosine similarity: 0.5428) and Gemma-3-4B-Instruct (ROUGE-L: 0.2024, cosine similarity: 0.5134) record the weakest quantitative performance. Notably, the strongest fine-tuned SLMs outperform all three commercial LLMs across most automatic metrics, where GPT-4.1 (ROUGE-L: 0.3038, cosine similarity: 0.6749), Gemini-2.5-Flash (ROUGE-L: 0.2771, cosine similarity: 0.6234) and Virtuoso-Large (ROUGE-L: 0.3161, cosine similarity: 0.6676) obtain lower scores, suggesting that domain-specific fine-tuning of SLMs improves response alignment for customer-service interactions.

In LLM-as-a-judge evaluation, GPT-4.1 achieves the strongest overall performance (4.146), while several fine-tuned SLMs show competitive performance, mainly in Human-Likeness and Tone and Clarity. LLaMA-3.1-8B-Instruct (3.794) outperforms Gemini-2.5-Flash (3.769), with strong Human-Likeness (4.115) and Tone and Clarity (4.149) (Table~\ref{tab:llm_judge_eval}). Qwen-3-8B-Instruct (3.743), LLaMA-3.2-3B-Instruct (3.718), Qwen-3-4B-Instruct (3.679) and Phi-4-Mini (3.619) also achieve solid overall performance. However, Continuity and Context Understanding and Task Appropriateness remain lower across SLMs, indicating limitations in maintaining dialogue coherence and task completion. Within the 3-4B parameter range, LLaMA-3.2-3B-Instruct, Qwen-3-4B-Instruct and Phi-4-Mini show stronger performance, while SmolLM3-3B (2.210) and Gemma-3-4B-Instruct (2.145) record the weakest results, especially in Continuity and Context Understanding. Within the less than 2B parameter range, Qwen-3-1.7B-Instruct (3.478) performs better than LLaMA-3.2-1B-Instruct (2.759), although both remain below 3-4B parameter SLMs. Stage-based analysis shows clear performance variation across conversation phases (Table~\ref{tab:llm_judge_stagewise}). In Early-stage interactions, LLaMA-3.1-8B-Instruct (3.819), LLaMA-3.2-3B-Instruct (3.800), Qwen-3-4B-Instruct (3.764) and Phi-4-Mini (3.700) achieve the strongest SLM results. In Mid-stage interactions, most SLMs show lower performance in Continuity and Task Appropriateness, indicating difficulty in maintaining coherent responses. However, LLaMA-3.1-8B-Instruct (3.737) and Qwen-3-8B-Instruct (3.712) remain competitive with Gemini-2.5-Flash (3.752). In Late-stage interactions, performance improves, where leading SLMs achieve scores above 4.1, with LLaMA-3.1-8B-Instruct (4.229) surpassing Gemini-2.5-Flash (3.818).

Human evaluation shows similar trends (Table~\ref{tab:human_eval}). Among the evaluated 3-4B parameter models, LLaMA-3.2-3B-Instruct achieves the highest SLM score (4.146), with strong performance in Human-Likeness (4.250), Continuity and Context Understanding (4.325) and Tone and Clarity (4.286). Qwen-3-4B-Instruct (4.069) and Phi-4-Mini (4.059) also perform at a similar level, though Task Appropriateness remains lower (3.553-3.721), indicating that accurate task resolution is still the main limitation despite strong fluency and contextual alignment. Compared to commercial LLMs, LLaMA-3.2-3B-Instruct (4.146) performs close to Gemini-2.5-Flash (4.191), with similar Human-Likeness and Tone and Clarity, while GPT-4.1 (4.627) and Virtuoso-Large (4.529) maintain higher scores in Continuity and Task Appropriateness. SmolLM3-3B (2.711) and Gemma-3-4B-Instruct (2.686) obtain lower scores across all qualitative dimensions, consistent with LLM-as-a-judge results, showing that not all SLM architectures within the 3-4B range are equally suited for this task. Stage-based human evaluation (Table~\ref{tab:human_eval_stagewise}) shows consistent variation across conversation phases. In Early-stage interactions, scores range from 3.882 to 4.005, with LLaMA-3.2-3B-Instruct (4.005) and Qwen-3-4B-Instruct (3.915) achieving the highest SLM results. Mid-stage performance remains stable (4.019-4.104) and competitive with Gemini-2.5-Flash (4.177), although GPT-4.1 (4.632) and Virtuoso-Large (4.517) remain higher. Late-stage performance is the strongest, where LLaMA-3.2-3B-Instruct (4.620), Phi-4-Mini (4.548) and Qwen-3-4B-Instruct (4.532) exceed Gemini-2.5-Flash (4.278), while GPT-4.1 (4.755) and Virtuoso-Large (4.753) remain above all SLMs.

Pairwise evaluation (Table~\ref{tab:pairwise_eval}) shows SLM strength mainly against Gemini-2.5-Flash, where Qwen-3-8B-Instruct achieves the highest SLM win rate (55.8\%), followed by LLaMA-3.1-8B-Instruct (52.9\%) and LLaMA-3.2-3B-Instruct (49.7\%). GPT-4.1 maintains clear dominance, with LLM win rates ranging from 54.2\% to 79.0\%, while the strongest SLM (Qwen-3-8B-Instruct) reaches 23.8\% wins. Against Virtuoso-Large, results are more balanced, with Qwen-3-8B-Instruct achieving 31.9\% wins, but LLM win rates (46.7-73.4\%) remain higher overall. Stage-wise pairwise results (Table~\ref{tab:pairwise_conversation_stages}) indicate that SLM competitiveness is most evident against Gemini-2.5-Flash, particularly in the mid stage, where Qwen-3-8B-Instruct achieves 60.45\% wins and LLaMA-3.1-8B-Instruct 55.97\%. Against GPT-4.1 and Virtuoso-Large, LLM win rates remain higher in early and mid stages across most models. Late-stage performance increases for several SLMs, with Phi-4-Mini (52.58\%), LLaMA-3.2-3B-Instruct (47.42\%) and Qwen-3-4B-Instruct (45.36\%) recording higher win rates against Gemini-2.5-Flash, while LLaMA-3.1-8B-Instruct achieves a 54.64\% combined win and tie rate against GPT-4.1.

Across all evaluation methods (Tables~\ref{tab:lexical_semantic_results}, 
\ref{tab:llm_judge_eval}, \ref{tab:human_eval}, \ref{tab:pairwise_eval}), clear performance differences are observed across models in the less than 2B, 3-4B and 8B parameter ranges. Within the less than 2B range, both models show limited competitiveness, with Qwen-3-1.7B-Instruct consistently outperforming LLaMA-3.2-1B-Instruct in LLM-as-a-judge (3.478 vs 2.759) and pairwise evaluation. Within the 3-4B range, two distinct performance groups are observed: LLaMA-3.2-3B-Instruct, Qwen-3-4B-Instruct and Phi-4-Mini demonstrate strong performance with competitive pairwise win rates against Gemini-2.5-Flash (49.7\%, 45.0\%, 43.9\%), while SmolLM3-3B and Gemma-3-4B-Instruct record substantially lower scores across all methods. In the 8B range, LLaMA-3.1-8B-Instruct and Qwen-3-8B-Instruct both show strong performance, with Qwen-3-8B-Instruct achieving the highest overall SLM pairwise win rate (55.8\% against Gemini-2.5-Flash). These results indicate that higher parameter capacity generally leads to stronger performance, although model architecture within the same parameter range also plays an important role.

SLM performance shows a consistent stage-wise pattern across all evaluation methods (Tables~\ref{tab:llm_judge_stagewise}, \ref{tab:human_eval_stagewise}, 
\ref{tab:pairwise_conversation_stages}), with moderate performance in the Early stage, lowest in the Mid stage and strongest in the Late stage. In the Early stage, top SLMs achieve competitive results, including LLaMA-3.1-8B-Instruct (3.819) in LLM-as-a-judge, LLaMA-3.2-3B-Instruct (4.005) in human evaluation and pairwise win rates reaching up to 44.44\% against Gemini-2.5-Flash. Mid-stage interactions are the most challenging, with the largest gaps in Continuity and Context Understanding and Task Appropriateness, although Qwen-3-8B-Instruct achieves a 60.45\% pairwise win rate against Gemini-2.5-Flash at this stage. In the Late stage, fine-tuned SLMs achieve their strongest results across all methods, where LLaMA-3.1-8B-Instruct surpasses Gemini-2.5-Flash in LLM-as-a-judge (4.229 vs 3.818), and LLaMA-3.2-3B-Instruct (4.620), Phi-4-Mini (4.548) and Qwen-3-4B-Instruct (4.532) exceed Gemini-2.5-Flash (4.278) in human evaluation, while Phi-4-Mini achieves a 52.58\% pairwise win rate, indicating stronger performance in resolution-focused interactions.

Results also show that model architecture impacts adaptability in customer-service QA settings, as this experiment is conducted under instruction-tuned settings. Although thinking budgets are set to enforce instruction-tuned behavior, Gemini-2.5-Flash and Qwen-3-8B-Instruct, which are reasoning-oriented models, show mixed performance when explicit reasoning is disabled through the removal of thinking budgets and tags. Under this setup, they perform lower than several instruction-focused SLMs. Qwen-3-8B-Instruct also underperforms compared to Qwen-3-4B-Instruct in some stages despite having more parameters (Table~\ref{tab:llm_judge_stagewise}), suggesting that reasoning-oriented architectures may require careful adaptation for instruction-tuned settings. In addition, Gemma-3-4B-Instruct, which is based on a multimodal architecture, also records lower performance across qualitative dimensions even after disabling multimodal components (Tables~\ref{tab:llm_judge_eval}, \ref{tab:human_eval}), indicating that such architectures may be less suitable for this task under instruction-tuned settings. Overall, fine-tuned SLMs, including 3-4B models such as Qwen-3-4B-Instruct, LLaMA-3.2-3B-Instruct and Phi-4-Mini, as well as 8B models such as LLaMA-3.1-8B-Instruct and Qwen-3-8B-Instruct, demonstrate strong performance for context-summarized multi-turn customer-service QA. However, SmolLM3-3B-Instruct and Gemma-3-4B-Instruct show consistent limitations across qualitative dimensions, indicating that not all SLM architectures are equally suitable for this task despite similar parameter sizes.

\section{Conclusion and Future Work}
\label{sec:conclusion}

This study provides a comprehensive evaluation of finetuned instruction-tuned SLMs for multi-turn context-summarized customer-service QA. In addition to the evaluation framework, we introduced a context-summarized synthetic multi-turn customer-service QA dataset designed to address privacy constraints and the lack of publicly available multi-turn conversational resources. Using automatic metrics, LLM-as-a-judge evaluation, human assessment, pairwise comparison and the conversation stage-based evaluation framework proposed in this work, the experiments examine how well SLMs maintain dialogue continuity, contextual understanding and response appropriateness across different phases of customer-service interactions. Results show that leading fine-tuned SLMs in the 3-8B parameter range, specifically Qwen-3-4B-Instruct, 
LLaMA-3.2-3B-Instruct, Phi-4-Mini, LLaMA-3.1-8B-Instruct and Qwen-3-8B-Instruct, have gained near-LLM performance in this task, particularly in human-likeness, tone and clarity and late-stage resolution. The proposed stage-based analysis demonstrates that SLM 
performance varies across Early, Mid and Late conversation phases, with mid-stage interactions presenting the greatest challenge due to higher contextual demands, and late-stage interactions producing the strongest results across all evaluation methods. Nevertheless, Continuity and Context Understanding and Task Appropriateness remain the primary limitations across all fine-tuned SLMs, with commercial LLMs such as GPT-4.1 and Virtuoso-Large continuing to achieve 
stronger results in these dimensions. SmolLM3-3B and Gemma-3-4B-Instruct show consistent limitations across all qualitative dimensions, indicating that model architecture plays a 
critical role in task suitability beyond parameter size alone.

These findings suggest that effective multi-turn customer-service systems do not necessarily require very large models, even though slight deviations remain between the best-performing SLMs and commercial LLMs. With context summarization and instruction tuning, SLMs offer a strong balance between performance and efficiency. The dataset contribution further supports reproducible research by providing structured multi-turn conversational instances suitable for evaluating dialogue continuity under privacy-aware conditions. This has practical and societal benefits, as smaller models reduce computational cost and energy usage, enabling wider access to customer-service automation while supporting privacy-conscious deployment. Overall, the study positions SLMs as a practical and scalable solution for multi-turn context-summarized customer-service QA, encouraging broader adoption of efficient conversational AI beyond high-resource settings.

Future work can extend evaluation to sectors such as healthcare, telecommunications and e-commerce to assess cross-domain robustness and the generalization capability of fine-tuned SLMs across different customer-service settings. Evaluation can also be expanded using real-world customer-service conversations, enabling assessment under more realistic interaction patterns and operational constraints. In addition, the dataset and evaluation can also be extended to multilingual settings, including code-mixed conversations, to assess the effectiveness of fine-tuned SLMs for multilingual customer-service interactions. Preference optimization methods, including Reinforcement Learning from Human Feedback (RLHF) and Reinforcement Learning from AI Feedback (RLAIF), can be explored to further improve conversational quality and better align SLMs with human expectations, particularly for dialogue continuity, tone control and task appropriateness \citep{wang2024comprehensivesurveyllmalignment}. Future work can also investigate end-to-end customer-service systems where both multi-turn context summarization and response generation are performed using SLMs, supporting on-premise deployment and strengthening privacy protection within customer-service QA systems. Benchmarking medium-scale models is also valuable, as results indicate that performance within SLMs generally increases with parameter count,
and further analysis can be conducted to examine the trade-offs between efficiency and conversational performance within small language models.

\subsection*{Limitations}

Since this study is conducted primarily on a banking-domain corpus, which may limit generalization to other customer-service domains. Although the synthetic dataset approach supports privacy preservation and avoids exposure of sensitive customer data, it may not capture the full variability of real-world interactions. These factors may influence the generalizability of the results beyond the evaluated setting. In addition, due to limited resources, human evaluation was conducted only on selected 3-4B SLMs and three commercial LLMs rather than across all evaluated models.

\subsection*{Acknowledgment}

We thank the human evaluators for their time and careful judgments during the qualitative evaluation process. This paper has been conducted in compliance with the ethical standards of the Informatics Institute of Technology (IIT).
We also acknowledge the Zame AI team for providing funding to support API usage, enabling large-scale model inference and evaluation.

\newpage
\appendix

\section*{Appendices}

\renewcommand{\thesection}{\Alph{section}}
\renewcommand{\appendixname}{Appendix}
\makeatletter
\renewcommand{\@seccntformat}[1]{\appendixname~\csname the#1\endcsname:~}
\makeatother

\section{Dataset statistics across splits}

\begin{table*}[!ht]
\centering
\small  % Added smaller font size
\setlength{\tabcolsep}{10pt}
\renewcommand{\arraystretch}{1.5}
\begin{tabular}{lrrrrrrr}
\hline
\textbf{Split} & \textbf{Samples} & \textbf{Total} & \textbf{Min} & \textbf{Avg} & \textbf{Max} & \textbf{Total} & \textbf{Avg} \\
 & & \textbf{Turns} & \textbf{Turns} & \textbf{Turns} & \textbf{Turns} & \textbf{Tokens} & \textbf{Tokens} \\
\hline
Train & 128,335 & 1,291,138 & 2 & 10.06 & 53 & 37,478,648 & 292.04 \\
Validation & 18,333 & 183,364 & 2 & 10.00 & 50 & 5,348,170 & 291.72 \\
Test & 36,669 & 368,650 & 2 & 10.05 & 58 & 10,714,768 & 292.20 \\
\hline
Whole Corpus & 183,337 & 1,843,152 & 2 & 10.05 & 58 & 53,541,586 & 292.02 \\
\hline
\end{tabular}
\caption{Dataset statistics across splits, including dialogue structure and token distribution. Token counts are computed using the GPT-4 tokenizer.}
\label{tab:dataset_stats}
\end{table*}

\section{Q-LoRA Finetuning Pipeline}

\vspace{10pt}
\begin{figure}[!htbp]
    \centering
    \includegraphics[width=\textwidth, height=10cm]{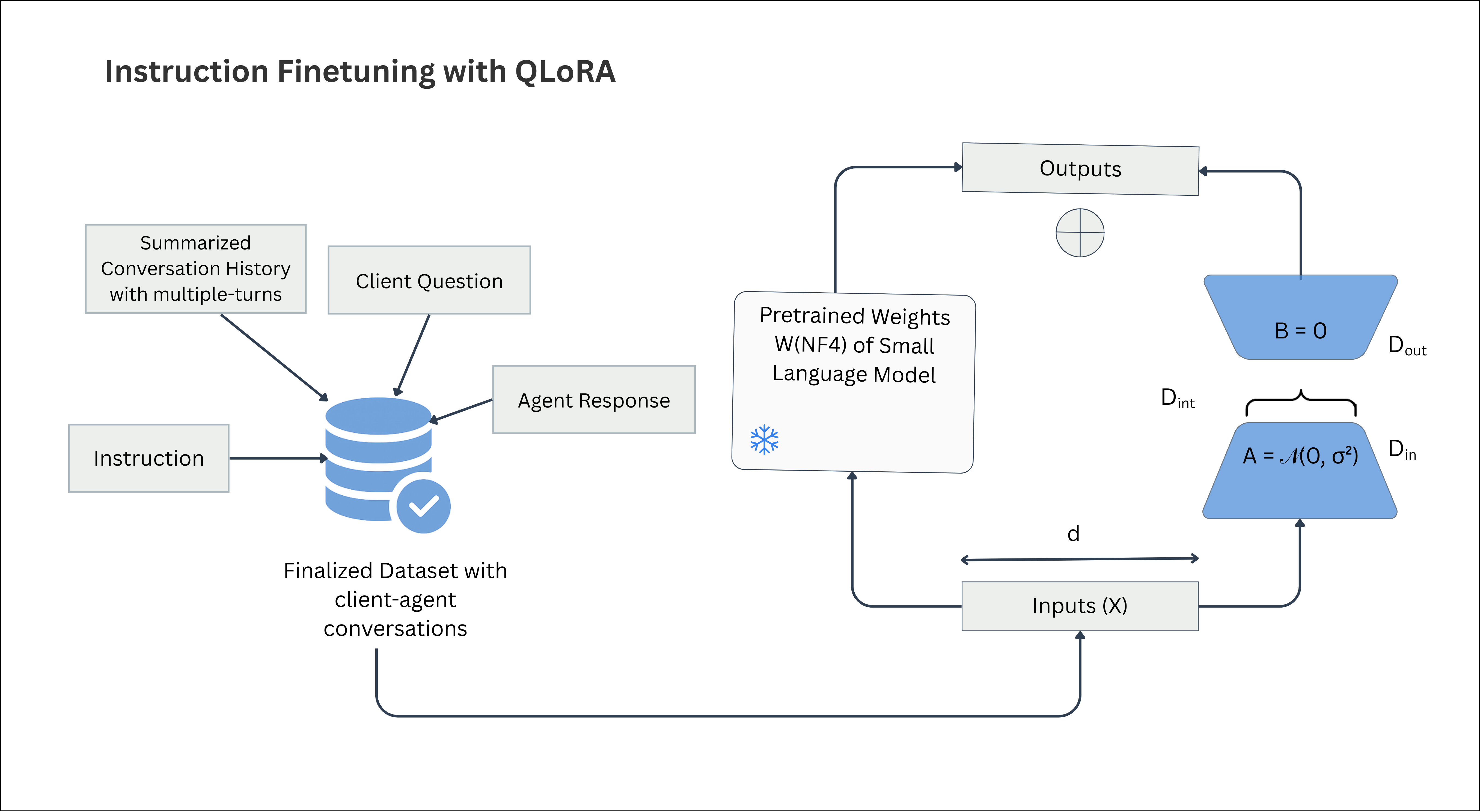}
    \caption{
    QLoRA based fine-tuning pipeline for context-summarized multi-turn customer-service QA.
    }
    \label{fig:context_summarized_example}
\end{figure}

\clearpage

\section{LLM-as-a-Judge Evaluation Prompt}

\label{app:llm_judge}

\small
The following LLM-as-a-judge prompt is used to evaluate generated responses for
context-summarized multi-turn customer-service QA. The same criteria are applied
to human evaluation.

\begin{tcolorbox}[
  colback=white,
  colframe=black!60,
  boxrule=0.4pt,
  arc=1pt,
  left=6pt,
  right=6pt,
  top=5pt,
  bottom=5pt
]

\lstset{
  basicstyle=\bfseries\ttfamily\fontsize{6pt}{6pt}\selectfont,
  breaklines=true,
  breakatwhitespace=false,
  breakindent=0pt,
  postbreak=\mbox{},
  columns=fullflexible,
  keepspaces=true,
  showstringspaces=false
}

\begin{lstlisting}
You are an expert evaluator specializing in customer-service interactions. Evaluate the Generated Response using the Client Question and Conversation History summary as context, along with a Reference Agent Response provided only as a high-quality example. The Reference Agent Response is provided only as guidance to illustrate what a professional, contextually appropriate answer might look like.The Generated Response should NOT replicate or closely mirror it.Instead, it should demonstrate human-like fluency, contextual understanding and professionalism while maintaining natural variation in expression and tone. Your task is to assess how human-like, contextually aware and professionally appropriate the Generated Response is.

Note:
The Conversation History Summary represents the main context that was used when generating the response.
The full Conversation History is provided only as additional background information to help you better understand the situation if needed.

Context Inputs:
Conversation History: {history}
Conversation Summarized History: {history_summary}
Client Question: {client_question}
Reference Agent Response (for guidance only): {ground_truth}
Generated Response: {generated_answer}

Evaluation Criteria and Scoring (1-5 each):

1. Human-Likeness:
This checks how natural and fluent the Generated Response sounds in normal spoken conversation.
It looks at flow, rhythm and how close it feels to real human speech.

Rating Scale:
1 = Highly robotic or unnatural
2 = Noticeably rigid or scripted
3 = Generally natural but somewhat stiff
4 = Natural and conversational
5 = Fully natural, smooth and human-like

2. Continuity and Context Understanding:
This evaluates how well the Generated Response integrates with the preceding conversation whether it maintains continuity,
references earlier details accurately and demonstrates awareness of situational context.

Rating Scale:
1 = Ignores or contradicts context
2 = Uses context incorrectly or inconsistently
3 = Uses context but with noticeable gaps
4 = Accurate and consistent use of context
5 = Fully coherent, precise integration of context

3. Tone and Clarity:
This measures verbal tone, emotional intelligence and clarity of expression.
It assesses professionalism, empathy, politeness and phrasing appropriateness for a spoken customer-service exchange.

Rating Scale:
1 = Unprofessional or unclear
2 = Understandable but flat or loosely structured
3 = Clear and appropriate, with standard professionalism
4 = Professional, well-structured and concise
5 = Highly polished, clear, respectful and well-balanced

4. Task Appropriateness:
This evaluates whether the Generated Response successfully and completely addresses the client's stated need,
while maintaining procedural accuracy typical of a service agent.

Rating Scale:
1 = Does not address the client's request
2 = Addresses request incompletely
3 = Provides an adequate response
4 = Fully addresses the request
5 = Fully addresses the request and adds meaningful helpful value

Return your answer as valid JSON only.
Do not include any explanation, commentary, additional text, or markdown formatting.
Output must contain JSON only and nothing else.All the below keys and there judgement score should be included in your json response. Strictly follow only below json output.always provide the score for all tasks in the json.

Output Format (return only JSON):
{{
  "Human-Likeness": [1-5],
  "Continuity-and-Context-Understanding": [1-5],
  "Tone-and-Clarity": [1-5],
  "Task-Appropriateness": [1-5]
}}
\end{lstlisting}

\end{tcolorbox}

\clearpage
\section{Pairwise Evaluation Prompt}

\label{app:pairwise}

\small
The following pairwise evaluation prompt is used to compare two generated
responses for the same context-summarized multi-turn customer-service QA.

\begin{tcolorbox}[
  colback=white,
  colframe=black,
  boxrule=0.6pt,
  arc=1.2pt,
  left=4pt,
  right=4pt,
  top=2pt,
  bottom=2pt
]

\lstset{
  basicstyle=\bfseries\ttfamily\fontsize{6pt}{6pt}\selectfont,
  breaklines=true,
  breakatwhitespace=false,
  breakindent=0pt,
  postbreak=\mbox{},
  columns=fullflexible,
  keepspaces=true,
  showstringspaces=false
}

\begin{lstlisting}
You are an expert evaluator specializing in customer-service interactions.

Your task is to compare two generated responses (Response A and Response B) to the same client query
and conversation context. Both responses were produced by different AI systems acting as
professional customer-service agents.

Use the Client Question and the Conversation History Summary as the main context for evaluation.
Use the full Conversation History only if additional background is needed.

A Reference Agent Response is provided only as an example of a good customer-service reply.
It is for general guidance only and should NOT be used as a comparison target.
Do not judge responses based on how similar they are to the reference.

Evaluation Criteria (all criteria are equally important):

1. Human-Likeness:
Which response sounds more natural and human-like in a real customer-service conversation?

2. Continuity and Context Understanding:
Which response better reflects the earlier conversation and correctly uses relevant details?

3. Tone and Clarity:
Which response is more professional, polite, empathetic and easy to understand?

4. Task Appropriateness:
Which response better addresses the client's request while following realistic customer-service practices?

If both responses are very similar in quality, choose the one that feels slightly more natural and
better connected to the context.

Context Inputs:
Conversation History: {history}
Conversation Summarized History: {history_summary}
Client Question: {client_question}
Reference Agent Response (example only): {ground_truth}

Response A:
{response_a}

Response B:
{response_b}

Choose the single better response overall.

Return your answer as valid JSON only.
Do not include any explanation or extra text.

Output Format:
{{ "winner": "A" or "B" }}
\end{lstlisting}
\end{tcolorbox}
\clearpage

\section{Context-Summarization Prompt}

\label{app:context_summary_prompt}

\small
The following prompt is used to generate context summaries by distilling prior multi-turn conversation history into a concise representation that preserves essential information, including the current status of the conversation.

\begin{tcolorbox}[
  colback=white,
  colframe=black,
  boxrule=0.6pt,
  arc=1.2pt,
  left=6pt,
  right=6pt,
  top=4pt,
  bottom=4pt
]

\lstset{
  basicstyle=\bfseries\ttfamily\fontsize{6pt}{6pt}\selectfont,
  breaklines=true,
  breakatwhitespace=false,
  breakindent=0pt,
  postbreak=\mbox{},
  columns=fullflexible,
  keepspaces=true,
  showstringspaces=false
}

\begin{lstlisting}
SUMMARY_PROMPT = """
You are a professional conversation summarization assistant.

Goal: Produce a clear, concise, factual summary of the conversation so far so that, the same customer service agent handling this client, can accurately answer their next question.

Include only information explicitly stated:
- Client’s issue/request and current status (who is the client and agent should be specifically mentioned if there names exist in the converstion)
- Verification steps completed or pending
- Exact names, accounts/identifiers, dates, amounts, and actions taken or agreed
- Commitments, deadlines, follow-ups
- Current status of the conversation

Exclude: greetings, filler, speculation, or invented details.

Style: neutral and professional. Vary sentence structure and phrasing to avoid repetition.

Output: one coherent detailed paragraph summary.

Conversation so far:
{conversation_data}
"""
\end{lstlisting}

\end{tcolorbox}

\section{Response Refinement Prompt}

\label{app:response_refinement}

\small
The following prompt is used to refine agent-generated responses so that they
sound natural, concise and appropriate for spoken customer-service interactions,
while preserving the original meaning and factual consistency.

\begin{tcolorbox}[
  colback=white,
  colframe=black,
  boxrule=0.6pt,
  arc=1.2pt,
  left=6pt,
  right=6pt,
  top=4pt,
  bottom=4pt
]

\lstset{
  basicstyle=\bfseries\ttfamily\fontsize{6pt}{6pt}\selectfont,
  breaklines=true,
  breakatwhitespace=false,
  breakindent=0pt,
  postbreak=\mbox{},
  columns=fullflexible,
  keepspaces=true,
  showstringspaces=false
}

\begin{lstlisting}
You are a call-center customer service agent. Rewrite the agent's answer so it
sounds like a real phone response that is clear, natural and helpful.

Requirements:
- Aim for 2-3 short spoken-style sentences. Use 4 only if a clear explanation is required.
- Start by acknowledging the customer's situation, then give the solution or next step right away.
- Keep facts consistent with the context; do not invent details.
- If key information is missing, ask just one short clarifying question and explain why it is needed.
- Avoid internal jargon, tool references, URLs, or repeating the full client question.
- Use greetings or sign-offs only when appropriate:
  * If the call is ending, close with a short thank-you using the company name.
  * If the issue is not resolved but can be completed now, offer to stay on the line until it is done.
- Keep the tone professional but natural, like a friendly call-center agent.
- Do not use emojis or symbols.
- Keep responses concise unless more detail is necessary for clarity.
- Use the client's name when possible.
- Only generate the final agent response; do not generate intermediate dialogue turns.
- Preserve the original meaning and supportive intent of the agent's answer.
- If the client asks for calculations, provide the result plainly without formulas.

Context:
Instruction: {instruction}
Previous client-agent summary:{summary}
Client's current question: {client_question}
Original agent answer: {agent_answer}

Rewrite only the final agent answer:
\end{lstlisting}

\end{tcolorbox}

\clearpage
\onecolumn

\section{Dataset Statistics and Distribution Analysis}

\label{app:dataset_stat}
\noindent
This presents token-length and dialogue structure statistics across the
train, validation and test splits, computed using the GPT-4 tokenizer.

\vspace{0.6em}

\begin{figure}[!htbp]
\centering
\begin{minipage}{\textwidth}
  \centering
  \includegraphics[width=\textwidth]{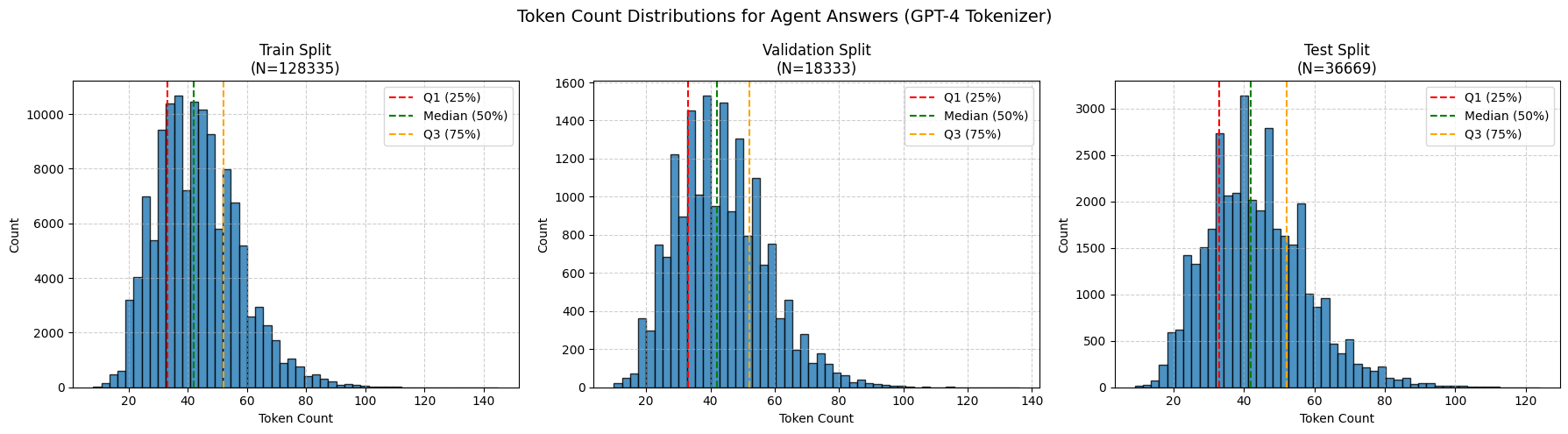}
  \caption{Token count distributions for agent answers across train, validation and test splits. Vertical lines indicate the first quartile (Q1), median and third quartile (Q3).}
  \label{fig:agent_answer_tokens}
\end{minipage}

\vspace{0.8em}

\begin{minipage}{\textwidth}
  \centering
  \includegraphics[width=\textwidth]{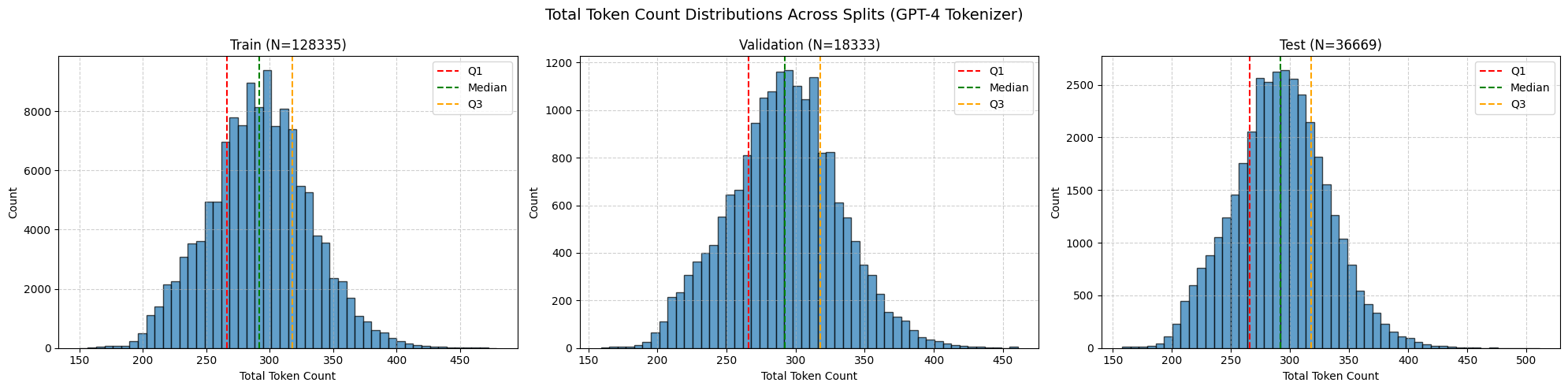}
  \caption{
Total token count distributions across dataset splits using the GPT-4 tokeniser, illustrating overall input length variability and quartile statistics computed over the combined \textit{instruction}, \textit{history\_summary}, \textit{client\_question} and \textit{refined\_agent\_answer} fields.   }
  \label{fig:total_token_counts}
\end{minipage}

\begin{minipage}{\textwidth}
  \centering
  \includegraphics[width=\textwidth]{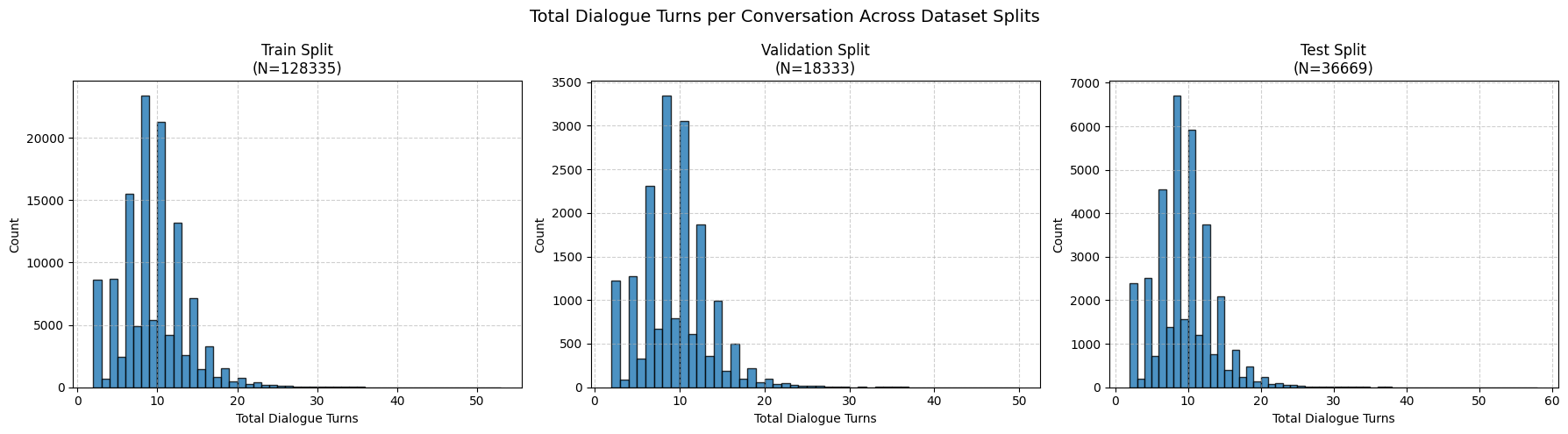}
  \caption{Distribution of total client-agent dialogue turns per conversation across train, validation and test splits.}
  \label{fig:dialogue_turns}
\end{minipage}

\end{figure}

\newpage

\section{Task-wise inter-evaluator agreement across evaluation dimensions}

\begin{figure*}[!htbp]
    \centering
    \includegraphics[width=12cm, height=9cm]{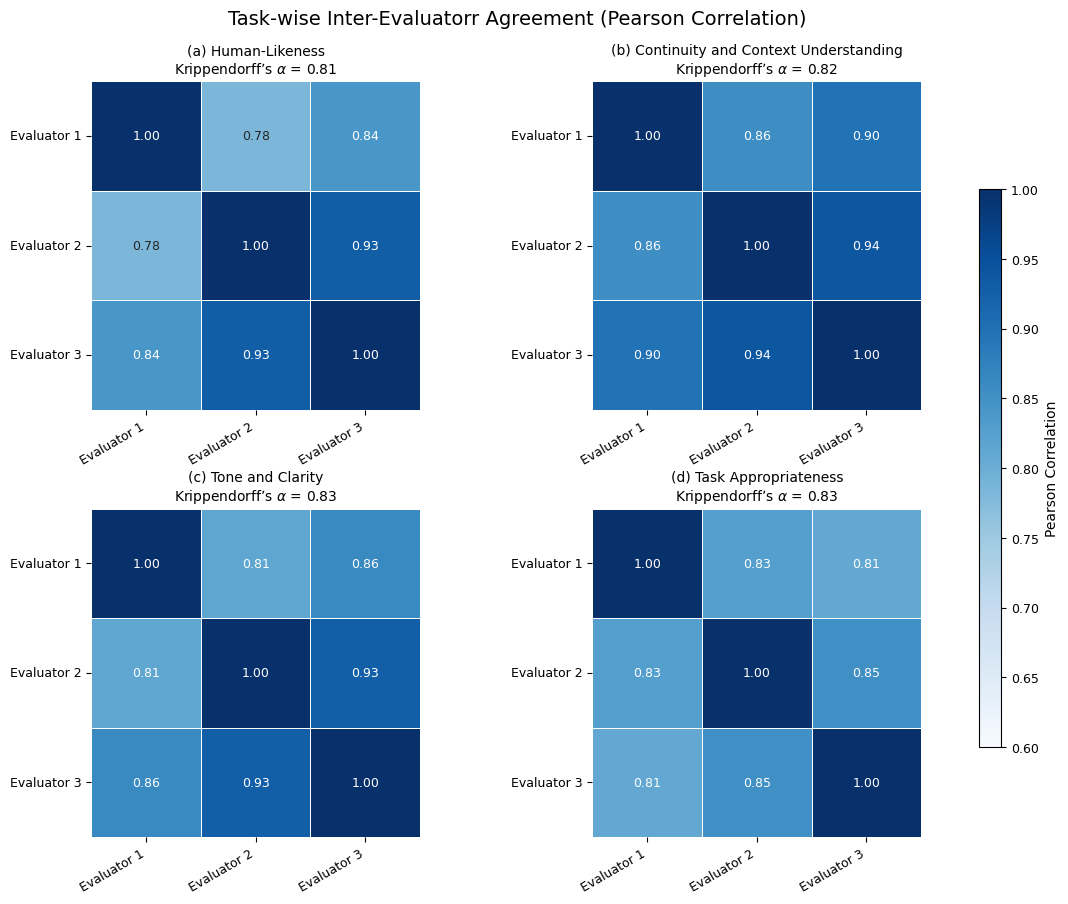}
    \caption{
    Each subfigure shows pairwise Pearson correlations between evaluators, with Krippendorff's $\alpha$ reported per dimension. Each evaluator assessed 500 responses per model (for all 3 -4B SLMs and the selected LLMs) using a 1-5 Likert scale. Strong agreement is observed across all criteria, indicating reliable human evaluation.
    }
    \label{fig:inter_evaluator_agreement}
\end{figure*}

\bibliographystyle{Frontiers-Harvard} %  Many Frontiers journals use the Harvard referencing system (Author-date), to find the style and resources for the journal you are submitting to: https://zendesk.frontiersin.org/hc/en-us/articles/360017860337-Frontiers-Reference-Styles-by-Journal. For Humanities and Social Sciences articles please include page numbers in the in-text citations 
\bibliography{test}

%%% Make sure to upload the bib file along with the tex file and PDF
%%% Please see the test.bib file for some examples of references

\end{document}